\documentclass[10pt,twocolumn,letterpaper]{article}

\usepackage{CVPR}              % To produce the CAMERA-READY version

\usepackage{times}
\usepackage{latexsym}
\usepackage{url}
\usepackage{inconsolata}
\usepackage{multirow}
\usepackage{graphicx}
\usepackage{color}
\usepackage{xcolor}
\usepackage{amsmath}
\usepackage{enumitem}
\usepackage{mdframed}
\usepackage{bm}
\usepackage{arydshln}
\usepackage{tcolorbox}
\usepackage{algorithm}
\usepackage{algorithmicx}
\usepackage[noend]{algpseudocode}
\usepackage{subcaption}
\usepackage{rotating}
\usepackage{booktabs}
\usepackage{marvosym}
\usepackage{footmisc}
\usepackage{amssymb}
\usepackage[utf8]{inputenc} 
\usepackage{colortbl}
\usepackage{pifont}
\usepackage{wrapfig}
\usepackage{array}
\usepackage{float}
\usepackage{stfloats}
\definecolor{darkgreen}{rgb}{0,0.8,0} 

\definecolor{cvprblue}{rgb}{0.21,0.49,0.74}
\usepackage[pagebackref,breaklinks,colorlinks,allcolors=cvprblue]{hyperref}

\def\paperID{12679} % *** Enter the Paper ID here
\def\confName{ICCV}
\def\confYear{2025}

\title{EscapeCraft: A 3D Room Escape Environment for Benchmarking Complex Multimodal Reasoning Ability}

\author{\textbf{Ziyue Wang}\textsuperscript{1\ \ding{171}}\footnotemark[1],
\textbf{Yurui Dong}\textsuperscript{3}\footnotemark[1], \textbf{Fuwen Luo}\textsuperscript{1}, \textbf{Minyuan Ruan}\textsuperscript{1}, \textbf{Zhili Cheng}\textsuperscript{1},\\
\textbf{Chi Chen}\textsuperscript{1}, \textbf{Peng Li}\textsuperscript{2\ \Letter}, \textbf{Yang Liu}\textsuperscript{1,2\ \Letter}\\ 
\textsuperscript{1}Dept. of Comp. Sci. \& Tech., Institute for AI, Tsinghua University, Beijing, China \\
\textsuperscript{2}Institute for AI Industry Research (AIR), Tsinghua University, Beijing, China \\
\textsuperscript{3}School of Management, Fudan University, Shanghai, China \\
}

\begin{document}
\twocolumn[{
\maketitle
\vspace{-35pt}
\begin{center}
    \centering
    \captionsetup{type=figure}
    \includegraphics[width=1.0\textwidth]{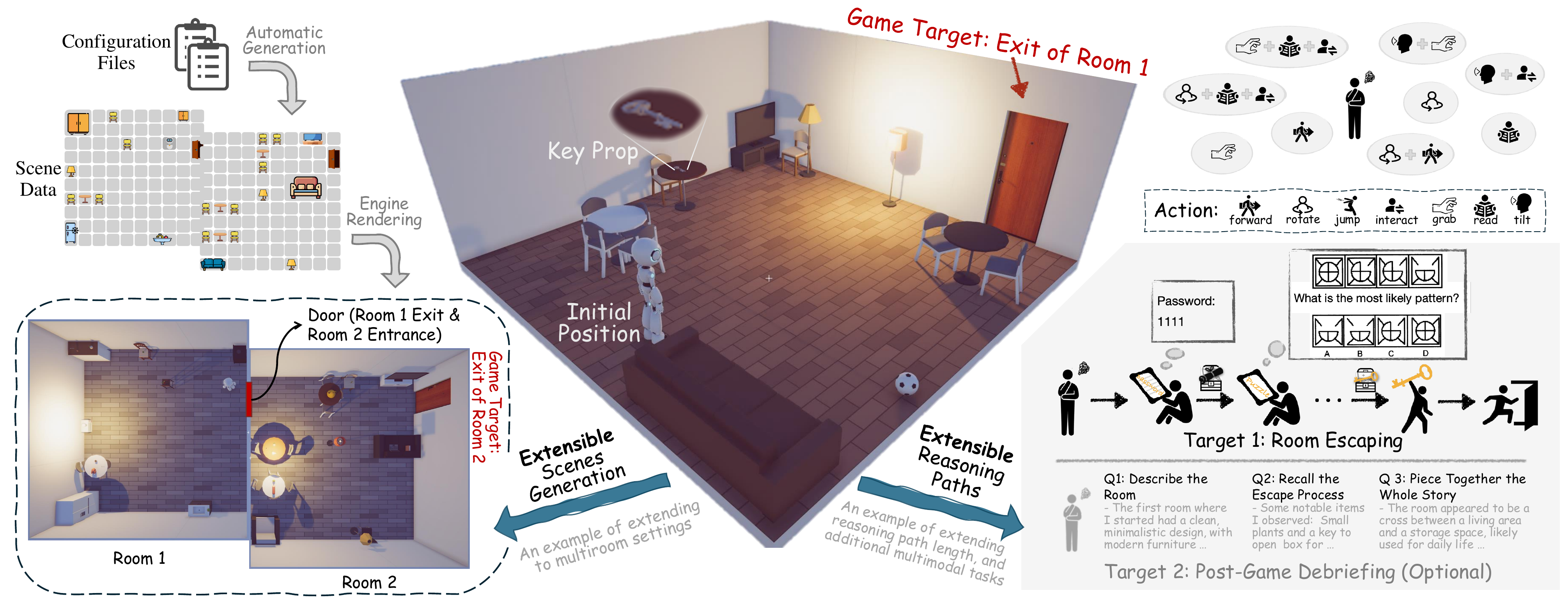}  
    \vspace{-20pt}
    \caption{Illustration of our proposed room escape environment EscapeCraft, which allows us to generate customized room scenes (left) and define groundtruth reasoning path of tasks (right). Based on EscapeCraft, we create MM-Escape benchmark, targeting at evaluating both the task completion performance and the entire multimodal reasoning process of MLLMs.}
  \label{fig:teaser}
\end{center}
}]

% \footnotetext[1]{Equal contribution, \textsuperscript{\ddag} Project lead, \textsuperscript{\Letter} Corresponding author}

\renewcommand{\thefootnote}{\fnsymbol{footnote}} 
    \footnotetext[1]{Equal contribution, \textsuperscript{\ding{171}} Project lead, \textsuperscript{\Letter} Corresponding author}
\renewcommand{\thefootnote}{\arabic{footnote}}

\vspace{-12pt}

\begin{abstract}
The rapid advancing of Multimodal Large Language Models (MLLMs) has spurred interest in complex multimodal reasoning tasks in the real-world and virtual environment, which require coordinating multiple abilities, including visual perception, visual reasoning, spatial awareness, and target deduction. 
However, existing evaluations primarily assess the final task completion, often degrading assessments to isolated abilities such as visual grounding and visual question answering. Less attention is given to comprehensively and quantitatively analyzing reasoning process in multimodal environments, which is crucial for understanding model behaviors and underlying reasoning mechanisms beyond merely task success.
To address this, we introduce MM-Escape, an extensible benchmark for investigating multimodal reasoning, inspired by real-world escape games. MM-Escape emphasizes intermediate model behaviors alongside final task completion. 
To achieve this, we develop EscapeCraft, a customizable and open environment that enables models to engage in free-form exploration for assessing multimodal reasoning. 
Extensive experiments show that MLLMs, regardless of scale, can successfully complete the simplest room escape tasks, with some exhibiting human-like exploration strategies. Yet, performance dramatically drops as task difficulty increases. Moreover, we observe that performance bottlenecks vary across models, revealing distinct failure modes and limitations in their multimodal reasoning abilities, such as repetitive trajectories without adaptive exploration, getting stuck in corners due to poor visual spatial awareness, and ineffective use of acquired props, such as the key. 
We hope our work sheds light on new challenges in multimodal reasoning, and uncovers potential improvements in MLLMs capabilities.
\footnote{GitHub repo: \url{https://github.com/THUNLP-MT/EscapeCraft}.}
\footnote{Home page: \url{https://thunlp-mt.github.io/EscapeCraft}.}
\end{abstract}    

\section{Introduction}
\label{sec:intro}

The rapid development of Large Language Models (LLMs) and Multimodal Large Language Models (MLLMs) have driven the advancement of diverse multimodal systems and applications for academic research~\cite{yan2025position, bhattacharya2024large}, industrial engineering~\cite{jose2024advancing}, and everyday assistance~\cite{wang2023gemini,fei2024multimodal}.
Multimodal reasoning is essential for these applications that require integrating multiple abilities such as visual perception, spatial awareness, and visual grounding~\cite{wang2024exploring}. For example, it enhances autonomous driving by improving the holistic understanding of multi-view information and localization, which are essential for vehicle actions and planning~\cite{ding2024holistic,fan2024mllmsulmultimodallargelanguage}, and also advances the general-purpose assistants in better performing visual and multimodal tasks in the wild~\cite{li2024multimodal, liu2024llava}.

Despite significant attention and effort towards improving multimodal reasoning abilities of MLLMs~\cite{wang2024exploring,lu2022learn,yang2023mm}, comprehensive evaluation remains underexplored for two key reasons. First, fundamental tasks such as visual grounding~\cite{yu2016modeling, yang2019fast,deng2021transvg} and image captioning~\cite{mokady2021clipcap,bai2018survey} are conducted in constraint environment and straightforward objectives, reducing the need for autonomous exploration. They primarily focus on identifying correct answers, such as bounding boxes and objects, without requiring coordinating multiple multimodal abilities.
Second, while multimodal tasks in open-world settings~\cite{ni2025peria,lin2021juewu,qin2024mp5} involve complex environments and objectives, they emphasize final task completion, often measured by success rate~\cite{Li_2024_CVPR}. This results in a lack of profound analysis over the reasoning process, leading to potentially inaccurate assessments of multimodal reasoning capabilities. Moreover, some open-world tasks provide structured knowledge libraries~\cite{ding2024minellama, wang2024rat} that standardize the reasoning mechanisms. These consequently limit the autonomy of models to conduct multimodal reasoning, making the reasoning more reliant on predefined knowledge rather than exploration of multimodal surroundings.

We argue that in open multimodal environment, including real-world settings and virtual simulators, complex multimodal reasoning should not be solely assessed by task completion results or isolated tasks. Instead, it is more practical and realistic to examine how models autonomously coordinate across multiple multimodal reasoning abilities. Recently, open environments such as Habitat~\cite{habitat19iccv}, AI2-THOR\cite{kolve2017ai2}, and OsWorld\cite{xie2025osworld} are widely discussed, where MLLMs are required to exhibit complex reasoning skills including visual searching, spatial understanding, tool utilization, and long-term decision-making. However, as summarize in Table~\ref{tab:compare_env}, there is a constraint on tasks or environments especially designed for evaluating multimodal reasoning in open-ended and interactive settings, leaving the evaluation of complex multimodal reasoning underestimated.

In this paper, we introduce \textbf{MM-Escape}, an extensible benchmark for evaluating complex multimodal reasoning, inspired by real-world escape games. To achieve this, we develop \textbf{EscapeCraft}, a customizable open environment that engages models in free-form exploration through the room escape task, assisting in comprehensive assessments of their multimodal reasoning abilities. 
MM-Escape provides solutions to the above issues of excessive concerns on task completion and simplified objectives, by emphasizing intermediate behaviors alongside final task completion. It measures the entire reasoning process consists of compound abilities.
While our designed games are intuitive for human players, they pose significant challenges for models regarding reasoning paths and interaction behaviors, exposing key limitations in current multimodal reasoning capabilities. 
As an additional bonus, we introduce post-game debriefing task, which is feasible only for the most challenging settings. It requires models to recall the escape process and reconstruct the story via collected clues, which assist in the future assessment of more complicated reasoning tasks.

We conduct comprehensive experiments and derive two key findings. First, although the overall performance remains far from human-level, recent MLLMs demonstrate strong multimodal reasoning ability. Notably, models like GPT-4o and Gemini-1.5-Pro achieve high escape rates and exhibit human-like exploration and reasoning strategies. While their interaction success rates are halved with steps more than doubled compared to human results. Second, as game difficulty increases, the model performance sharply declines, revealing distinct failure modes across models. For instance, GPT-4o tends to repeat actions and get trapped in historical trajectories when reasoning paths grow longer, Gemini struggles with spatial awareness and often gets stuck in corners. 
We also thoroughly discuss post-game debriefing and other extensible settings in Section~\ref{label:ana}. These findings highlight the need for greater attentions toward comprehensive analysis and improvement of multimodal reasoning abilities.
Our contributions are as follows:
\begin{itemize}
    \item We introduce a benchmark, MM-Escape, to advance comprehensive evaluation of multimodal reasoning for MLLMs, by quantitatively evaluating intermediate reasoning process alongside task completion performance.
    \item Our benchmark features free-form exploration, requiring models to autonomously coordinate multiple multimodal reasoning abilities in the multimodal room escape task.
    \item We thoroughly investigate model behaviors using MM-Escape and identify distinct limitations across models. Our analysis provides detailed insights, highlighting future optimization and potential real-world applications.
\end{itemize}
\begin{table*}[t]
    \centering\scriptsize
    % \resizebox{\textwidth}{!}{
    \begin{tabular}{l|c|c|c|c|c|c}
    \toprule
        Benchmark & Scenario & Task & Multimodal & Environment Type & URP & PA \\
    \midrule
        TextWorld~\citep{cote2019textworld} & Text Game & Simplified Text Games & \textcolor{red}{\ding{55}} &Open Environment & \textcolor{darkgreen}{\ding{51}} & \textcolor{red}{\ding{55}} \\
        EscapeBench~\citep{qian2024escapebenchpushinglanguagemodels} & Text Game & Room Escape & \textcolor{red}{\ding{55}} & Open Environment & \textcolor{darkgreen}{\ding{51}} & \textcolor{darkgreen}{\ding{51}}\\
    \midrule
        OpenEQA~\citep{majumdar2024openeqa} & Video & Ego-centric QA & \textcolor{darkgreen}{\ding{51}} & Controlled Environment & \textcolor{red}{\ding{55}} & \textcolor{red}{\ding{55}} \\
        EgoSchema~\citep{mangalam2023egoschema} & Video & Ego-centric QA & \textcolor{darkgreen}{\ding{51}} & Controlled Environment & \textcolor{red}{\ding{55}} & \textcolor{red}{\ding{55}} \\
        VSI-Bench~\citep{yang2024thinking} & Video & Visual-Spatial QA & \textcolor{darkgreen}{\ding{51}} & Controlled Environment & \textcolor{red}{\ding{55}} & \textcolor{red}{\ding{55}} \\
        MineRL~\citep{guss2019minerl} & Video Game & Goals in Minecraft & \textcolor{darkgreen}{\ding{51}} & Open Environment & \textcolor{red}{\ding{55}} & \textcolor{red}{\ding{55}} \\
        OSWorld~\citep{xie2025osworld} & OS Environment & Computer Use & \textcolor{darkgreen}{\ding{51}} & Open Environment & \textcolor{red}{\ding{55}} & \textcolor{red}{\ding{55}} \\
        ALFRED~\citep{shridhar2020alfred} & Embodied Environment & Household Task & \textcolor{darkgreen}{\ding{51}} & Open Environment & \textcolor{red}{\ding{55}} & \textcolor{red}{\ding{55}} \\
        EmbodiedEval~\citep{cheng2025embodiedeval} & Embodied Environment & Household Task & \textcolor{darkgreen}{\ding{51}} & Open Environment & \textcolor{red}{\ding{55}} & \textcolor{red}{\ding{55}} \\
    \midrule
        MM-Escape (Ours) & Embodied Environment & Room Escape & \textcolor{darkgreen}{\ding{51}} & Open Environment & \textcolor{darkgreen}{\ding{51}} & \textcolor{darkgreen}{\ding{51}} \\
    \bottomrule
    \end{tabular}
    % }
    \vspace{-9pt}
    \caption{Benchmarks aiming at evaluating complex reasoning ability of models. \textbf{URP} means \textit{unpredictable reasoning path}. \textbf{PA} means \textit{process assessment}. \label{tab:compare_env}}
    \vspace{-15pt}
\end{table*}

\section{Related Work}
 
\subsection{Complex Reasoning Abilities of MLLMs}
Recent research on MLLMs has moved beyond addressing simple tasks, such as image captioning and image retrieval, and instead focuses on enhancing model abilities towards more complex tasks and more realistic scenarios. For example, MLLMs can operate digital devices such as computers~\citep{yang2024swe,cua2025computeragent} and mobile phones~\cite{wang2025mobileagentvlearningmobiledevice,wang2025mobile}, play video games requiring long action sequences~\citep{chen2024vlmsplayactionroleplaying,de2024willgpt4rundoom}, and perform robotic manipulation in the real world~\citep{driess2023palme,zhu2024retrieval}. Unlike tasks such as visual question answering and visual grounding, which have standardized answers, these complex tasks allow diverse solutions and engage in open environments. As a result, multimodal reasoning ability is becoming increasingly important for achieving more advanced targets.

\subsection{Complex Multimodal Reasoning Evaluation}
Researchers have made efforts to evaluate the visual reasoning ability of MLLMs in complex tasks which require models to coordinate multiple multimodal abilities~\cite{mangalam2023egoschema,majumdar2024openeqa,yang2024thinking}. However, many of them leverage videos with predefined perspectives and trajectories as input. Models cannot interact with environments by themselves.

Some other works focus on complex open worlds, such as Minecraft~\citep{wang2023voyager,wang2023describe,qin2024mp5,wang2025jarvis,lin2021juewu,li2024optimus}, to evaluate model abilities of multimodal reasoning. However, two issues still exist. First, due to clearly defined game rules, standardized guides are available from websites like Minecraft Wiki~\footnote{\url{https://minecraft.wiki/}}, and training datasets of game demonstrations~\citep{guss2019minerl}, making it feasible for models to reason based on their commonsense or relying on external knowledge rather than observations during free-form exploration. Second, evaluations are often outcome-oriented, for example, focusing on specific goals like obtaining a diamond in Minecraft, while how models explore the open world is ignored. Similar issues exist in other complex open-ended tasks in open worlds like computer use~\citep{xie2025osworld} and robotic manipulation~\citep{shridhar2020alfred,yang2025embodiedbench}.

Compared to existing benchmarks, our work features in three aspects: a) we support sufficient interactions between models and environments; b) environments can be flexibly configured, preventing models from solely relying external knowledge to achieve the goals; c) reasoning processes are also evaluated alongside final results. These enable us to understand model abilities more comprehensively.

\vspace{-3pt}
\subsection{Multimodal v.s. Pure Text Environments}

Previous works investigated reasoning ability of models in pure text scenarios by allowing models to interact with environments via text interfaces~\citep{cote2019textworld,xu2023exploring,qian2024escapebenchpushinglanguagemodels}. However, some multimodal information cannot be easily and precisely expressed in verbal forms, hindering reasoning ability based on multimodal information to be evaluated in pure text environments. Additionally, models enjoy enhanced decision-making autonomy within multimodal environments compared to text environments. For example, models may decide their rotation angles or travel distances based on multimodal inputs when requiring to observe target objects or nearby surroundings, but it is hard to make such decisions based on text descriptions. 
Therefore, we believe that multimodal environments are essential for evaluating reasoning abilities of models, and propose EscapeCraft and MM-Escape to address the above issues.
\begin{figure*}
    \centering
    \includegraphics[width=.99\linewidth]{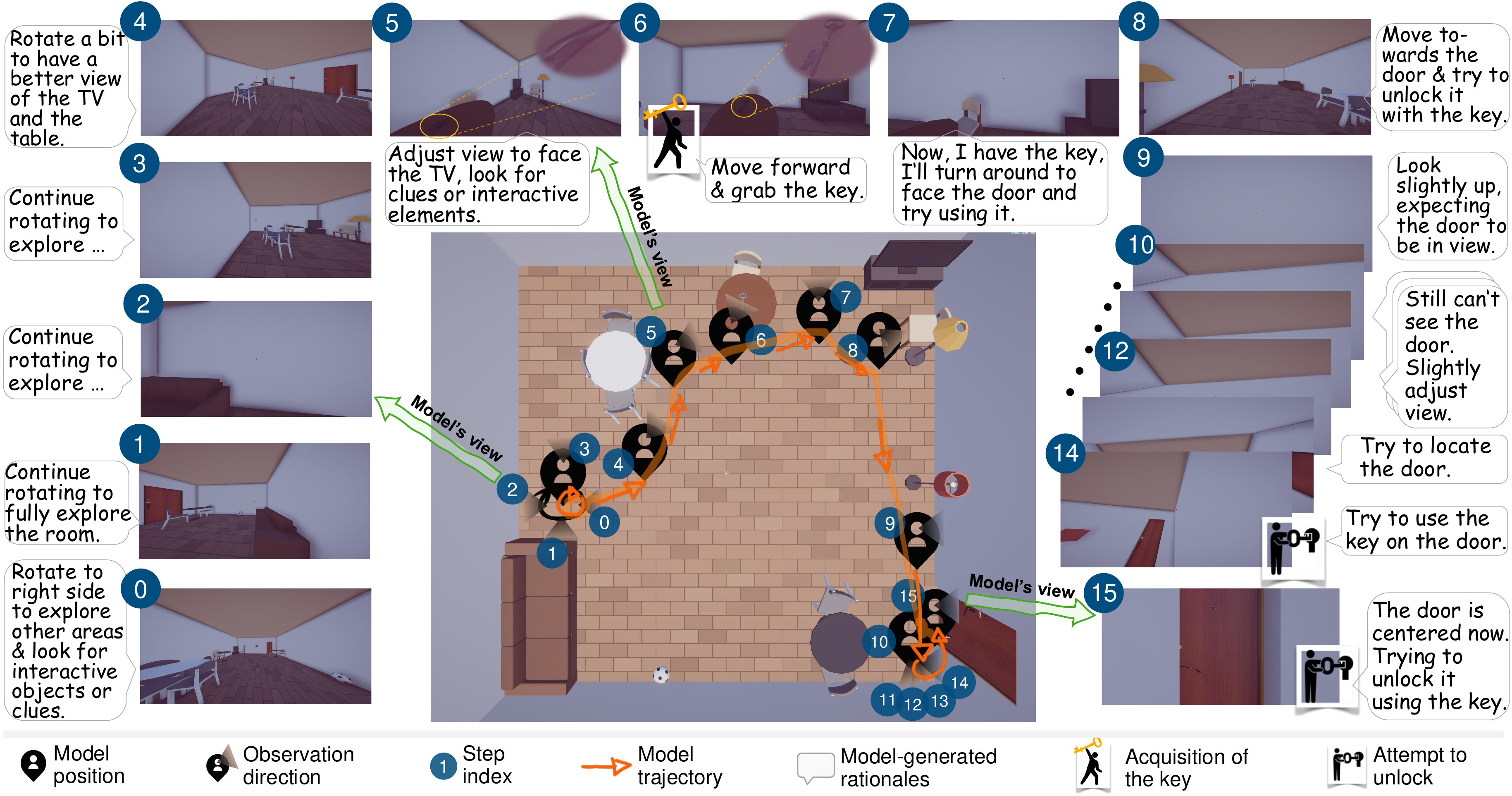}
    \vspace{-10pt}
    \caption{An example of success escape path from Gemini-1.5-pro. For the first four steps, it rotates, without moving, to observe the surroundings, and then starts moving towards the direct on the observed tables in the following steps. For step 5 and 6, we zoom in to the key to make it clearer. The key is distorted because the view of model is tilting downwards to observe objects on the table. From step 10 to 14, the model is already close to the door, and is slowly rotating and tilting to locate the door.}\label{fig:example}
    \vspace{-10pt}
\end{figure*}

\section{MM-Escape}
In this paper, we address comprehensive investigation of complex multimodal reasoning for MLLMs by introducing \textbf{MM-Escape}, an extensible benchmark. To support this, we design \textbf{EscapeCraft}, a highly customizable and open environment that engages models in free-form exploration for quantitatively assessing the intermediate multimodal reasoning process, as depicted in Figure~\ref{fig:teaser}. We highlight that complex multimodal reasoning requires an autonomous integration of multiple multimodal abilities, and that evaluating the reasoning process is as crucial as assessing final task completion. We will first describe our introduced tasks in the following section, and detail the construction of our benchmark along with its corresponding environment.

\subsection{Task Definition}

Inspired by real-world game, room escape, we propose to place MLLMs in an escape escape scenario to investigate their multimodal reasoning abilities in an open-environment, where models should autonomously explore multimodal surroundings, and reason toward both temporary and long-term goals. We name the introduced task, multimodal room escape. It consists of a mandatory target, the \textit{room escaping} task, aiming at escaping a locked room, and an optional target the \textit{post-game debriefing} task, requiring to reconstruct the story discovered during the escaping. Our multimodal room escape task assesses the entire reasoning process rather than solely focusing on final game completion.

\vspace{-12pt}
\paragraph{Room Escaping Task.} This task presents a ultimate goal of exiting the room, as shown in Figure~\ref{fig:teaser} (Game Target) and a detailed example is demonstrated in Figure~\ref{fig:example}. It requires models to fully explore and interact with the multimodal environment, search for props and clues, identify the exit, and correctly use props to unlock the door. 
We do not exploit step-by-step instructions to restricted model actions, ensuring them to freely explore the environment without constraints, and automatically deduce current or short-term goals.
This task evaluates the integrated multimodal reasoning ability, including object recognition, visual search, visual reasoning, target deduction, spatial reasoning, and prop utilization. Please refer to Supplementary Materials Sec.~\ref{app:task} for detailed discussion over full required abilities.

Post-game debriefing task differs from the room escaping task that involves reasoning about past experiences, current states, and future plans, as it takes place after the game is completed\footnote{This task is also common in real-world escape game.}. It serves as an optional tasks because its preliminary requirement is to collect all props in the game and exit the room successfully, which is quite challenging for most of current models. Therefore, for a fair comparison, we only apply this target to cases where models correctly complete the game as shown in Figure~\ref{fig:teaser}. It emphasizes logicality and consistency of reconstruction, requiring models to reflect on the past experiences and events.

\subsection{Construction and Design of Environment}
We develop EscapeCraft to place models in an escape game, a free-form exploration environment, to comprehensively investigate complex multimodal reasoning ability of models. It supports customizable and extensible scene generation for our escape game with minimal manual effort.

\vspace{-9pt}
\paragraph{Room Scene Generation} To support room escape task, we generate diverse and interactable room environments, which requires efficient and large-scale scene data generation and rendering. We develop an environment, EscapeCraft, by extending ProcTHOR~\cite{deitke2022} and Legent~\cite{cheng2024legent}, which are originally labor-intensive regarding the scene generation process. We enable automatic size adaptation to predefined or customized configurations, such as room scale, number of rooms, and required furniture, by incorporating 3D furniture models with annotated size information. Critical objects are made fully interactable to serve as props and clues to assist in completing the task. The flexibility and interactivity of objects allow for automatic large-scale 3D room generation following the requirement of our benchmark. EscapeCraft is highly customizable and extensible, making it well-suited for the room escape task. Detailed construction can be found in Supplementary Material~\ref{app:env_cons}.

\vspace{-12pt}
\paragraph{Action Space} We define three types of actions, moving, view adjustment, and interaction. The moving action, i.e. moving forward, allows the model to change its position to perceive objects at different depth. View adjustment enables perception from different angles and facilitates object selection for interaction, including horizontal or vertical rotation, and looking at specific coordinates. Interaction actions contain grabbing, using, reading, and inputting, allowing  models to obtain and utilize props from the environment, and process messages displayed by the props. These actions can be executed individually or integratedly in a multi-action way. See Supplementary Materials Sec.~\ref{app:action} for details.

\vspace{-12pt}
\paragraph{Inventory system} To enable model players to acquire and utilize props within the environment, we design an inventory system that allows models to store and manage acquired items, access detailed information about them, and use them as needed. This system assists models in effectively using props, and successfully escaping the room.

\subsection{MM-Escape Benchmark}

\begin{figure}
    \centering
    \includegraphics[height=4cm]{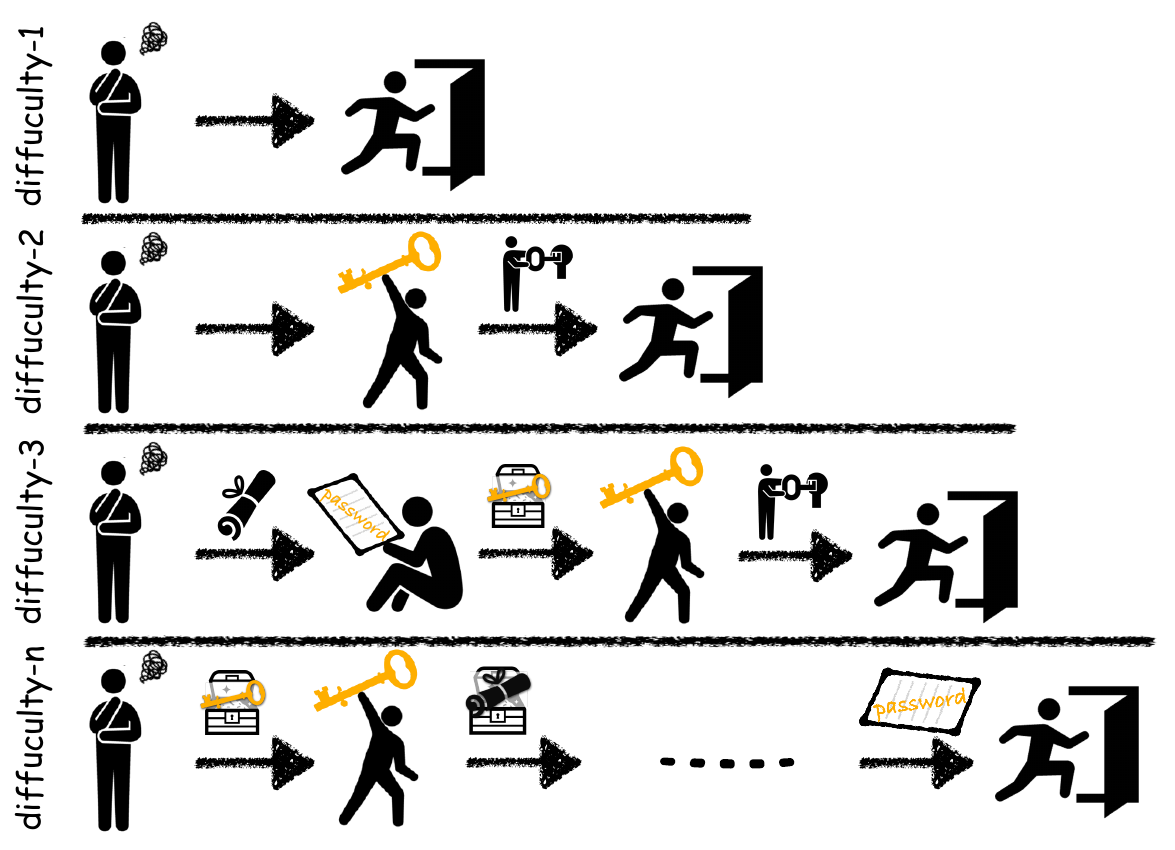}
    \vspace{-8pt}
    \caption{Illustration of difficulties, showing required paths for the reasoning process of successful escapes regarding each levels. The levels can be customized and extended as depicted by ``difficulty-n'' example, and as demonstrated in Figure~\ref{fig:teaser} (lower right part).}
    \label{fig:levels}
    \vspace{-14pt}
\end{figure}

\paragraph{Game Settings.\label{sec:game_setting}} 
We introduce an automatic reasoning chain generation procedure by configuring the \textit{Prop Chain}. In detail, this is a singly linked list representing the ordered sequence of items and interactions required to complete the game. Each node in the chain corresponds to an interactive element, such as a key, a locked box, or a note with a password, where the tail node represents the exit point of the game. To construct a complete escape game setting, we annotate the links between nodes in the prop chain to define the ways to obtain different props (such as unlimited acquisition or requiring a key to open, etc.) and their inclusion relationships (for example, a key can be placed in a box). See Supplementary Material Sec.~\ref{app:data_cons} for details.

Following this, we employ varying difficulty levels in MM-Escape to facilitate in-depth assessments of complex multimodal reasoning. Shown in Figure~\ref{fig:levels}, difficulty is primarily determined by the predefined prop chain of a game, where longer chains correspond to higher difficulty.  We define three standard difficulty levels for individual rooms: 
\begin{itemize}
    \item Difficulty-1: The simplest one-hop reasoning path where no props are needed to unlock the door. Models can exit by locating the door and interacting with it directly.
    \item Difficulty-2: A two-hop reasoning path requiring an additional key or password compared to Difficulty-1. Models should search for the key or password and interact with it to unlock the door.
    \item Difficulty-3: A three-hop reasoning path requiring both a password and a key, with one additional hop to Difficulty-2. This level challenges models with spatial reasoning, visual search, and prop utilization.
\end{itemize}
Since the prop chain can grow infinitely, our difficulty levels are inherently extendable. Moerover, the type of questions or tasks in each reasoning hop are customizable and interchangeable, further enhancing the difficulty and flexibility of MM-Escape. We also explore some extended settings that incorporate with other tasks, such as embodied QA and visual logical reasoning, with case studies in Section~\ref{sec:case_study}

Additionally, to further investigate the behavior and complex multimodal reasoning abilities of MLLMs, we introduce a multi-room setting by combining two standard single rooms.
We create multiple multi-room combinations, each containing two individual rooms. The configurations include: two Difficulty-1 rooms, two Difficulty-2 rooms, and a mixed setting of Difficulty-1 and Difficulty-2. In the multi-room setting, models start in the first room, which has only one exit. Upon successfully exiting, models enter the second room and search for the final game exit. However, this setting presents a greater challenge than the single-room scenario, as there are two doors in the second room, requiring models to distinguish between exit and entrance based on their corresponding surroundings. 

\vspace{-12pt}
\paragraph{Statistics of MM-Escape}
For individual room settings, we generated 11 scenes for each of Difficulty-1 and Difficulty-2, and 21 scenes for Difficulty-3. As there are two types of props required by Difficulty-3, we enable a key-first and a password-first prop chains. 
For multi-room settings, we generated 10 scenes for the three different combinations introduced above.
There are totally 63 scenes for standard evaluation of our benchmark, which could be further extended for future research without adapting the environment.
These scenes are categorized into four distinct styles: living room (14), kitchen (19), bathroom (19), and bedroom (11). And different objects are automatically placed within each scene to correspond to its specific style. On average, difficulty-1 scenes contain 20.18 objects, difficulty-2 contains 14.55, and difficulty-3 contains 15.24. The objects are logically arranged in a manner consistent with real-life settings and randomly distributed within the rooms to serve as part of the tasks for models.

\begin{table*}[t]
    \centering
    \resizebox{\textwidth}{!}{
        \begin{tabular}{@{\hspace{0.1cm}}l|c@{\hspace{0.4cm}}ccc|c@{\hspace{0.3cm}}cccc|c@{\hspace{0.3cm}}cccc|c@{\hspace{0.1cm}}}
        \toprule 
         \multirow{2}{*}[-1.5ex]{Models} & \multicolumn{4}{c|}{Difficulty-1} & \multicolumn{5}{c|}{Difficulty-2} & \multicolumn{5}{c|}{Difficulty-3} & \multirow{2}{*}[-1.5ex]{{\begin{tabular}[c]{@{}c@{}}AVG\\ER (\%)$\uparrow$\end{tabular}}} \\ \cmidrule(lr){2-5} \cmidrule(lr){6-10}\cmidrule(lr){11-15} 
         & {\begin{tabular}[c]{@{}c@{}}ER\\(\%)$\uparrow$\end{tabular}} & Steps$\downarrow$ & {\begin{tabular}[c]{@{}c@{}}Grab \\SR (\%)$\uparrow$\end{tabular}} & {\begin{tabular}[c]{@{}c@{}}Grab\\Ratio\end{tabular}} & 
         {\begin{tabular}[c]{@{}c@{}}ER\\(\%)$\uparrow$\end{tabular}} & {\begin{tabular}[c]{@{}c@{}}Prop\\(\%)$\uparrow$\end{tabular}} & Steps$\downarrow$ &  {\begin{tabular}[c]{@{}c@{}}Grab \\SR (\%)$\uparrow$\end{tabular}} & {\begin{tabular}[c]{@{}c@{}}Grab\\Ratio\end{tabular}} & 
         {\begin{tabular}[c]{@{}c@{}}ER\\(\%)$\uparrow$\end{tabular}} & {\begin{tabular}[c]{@{}c@{}}Prop\\(\%)$\uparrow$\end{tabular}} & Steps$\downarrow$ &  {\begin{tabular}[c]{@{}c@{}}Grab \\SR (\%)$\uparrow$\end{tabular}} & {\begin{tabular}[c]{@{}c@{}}Grab\\Ratio\end{tabular}} & \\
         \midrule 
         Human & 100.00 & \phantom{0}5.73 & 95.45 & 0.19 & 100.00 & 100.00 & 13.64 & 81.81 & 0.19 & 100.00 & 100.00 & 21.45 & 75.45 & 0.19 & 100.00 \\
         \midrule
         GPT-4o & \cellcolor{green!20}{\textbf{100.00}} & \cellcolor{green!20}{\textbf{11.27}} & \cellcolor{green!20}{37.82} & \cellcolor{green!20}{0.42} & 
         \cellcolor{yellow!20}{\phantom{0}\textbf{72.73}} & \cellcolor{yellow!20}{\phantom{0}\underline{81.82}} & \cellcolor{yellow!20}{\textbf{36.73}}  & \cellcolor{yellow!20}{\textbf{36.73}} & \cellcolor{yellow!20}{0.26} &
         \cellcolor{orange!20}{\phantom{0}\textbf{71.36}} & \cellcolor{orange!20}{\phantom{0}\textbf{90.00}} & \cellcolor{orange!20}{\phantom{0}\textbf{50.19}} &  \cellcolor{orange!20}{\textbf{31.36}} & \cellcolor{orange!20}{0.35} & \phantom{0}81.36 \\
         Gemini-1.5-pro & \cellcolor{green!20}{\phantom{0}81.82} & \cellcolor{green!20}{21.18} & \cellcolor{green!20}{\textbf{49.18}} & \cellcolor{green!20}{0.39} & 
         \cellcolor{yellow!20}{\phantom{0}\underline{54.55}} & \cellcolor{yellow!20}{\phantom{0}\textbf{90.91}} & \cellcolor{yellow!20}{\underline{47.82}} & \cellcolor{yellow!20}{14.89}  & \cellcolor{yellow!20}{0.44} &
         \cellcolor{orange!20}{\phantom{0}\underline{46.82}} & \cellcolor{orange!20}{\phantom{0}\underline{74.49}} & \cellcolor{orange!20}{\phantom{0}\underline{73.18}} & \cellcolor{orange!20}{10.43} & \cellcolor{orange!20}{0.48} & \phantom{0}61.06 \\
         Claude 3.5 Sonnet & \cellcolor{green!20}{\phantom{0}72.73} & \cellcolor{green!20}{22.09} & \cellcolor{green!20}{30.64} & \cellcolor{green!20}{0.36} & 
         \cellcolor{yellow!20}{\phantom{0}45.45} & \cellcolor{yellow!20}{\phantom{0}54.55} & \cellcolor{yellow!20}{57.45} & \cellcolor{yellow!20}{\underline{20.64}} & \cellcolor{yellow!20}{0.17} & 
         \cellcolor{orange!20}{\phantom{0}39.61} & \cellcolor{orange!20}{\phantom{0}54.83} & \cellcolor{orange!20}{\phantom{0}82.36} & \cellcolor{orange!20}{\underline{16.21}} & \cellcolor{orange!20}{0.22} & \phantom{0}52.60 \\
         Doubao 1.5 Pro & \cellcolor{green!20}{\phantom{0}\underline{91.91}} & \cellcolor{green!20}{\underline{16.27}} & \cellcolor{green!20}{\underline{44.68}} & \cellcolor{green!20}{0.27} & 
         \cellcolor{yellow!20}{\phantom{0}45.45} & \cellcolor{yellow!20}{\phantom{0}54.55} & \cellcolor{yellow!20}{63.18} & \cellcolor{yellow!20}{13.63} & \cellcolor{yellow!20}{0.25} & 
         \cellcolor{orange!20}{\phantom{00}9.52} & \cellcolor{orange!20}{\phantom{0}33.33} & \cellcolor{orange!20}{\phantom{0}93.19} & \cellcolor{orange!20}{\phantom{0}6.76} & \cellcolor{orange!20}{0.26} & \phantom{0}48.96 \\         
         % GLM-4v plus \\
         % MiniCPM-v 2.6 \\
         Llama-3.2-11b-vision & \cellcolor{green!20}{\phantom{0}63.64} & \cellcolor{green!20}{23.55} & \cellcolor{green!20}{31.36}  & \cellcolor{green!20}{0.35}  & 
         \cellcolor{yellow!20}{\phantom{00}0.00} & \cellcolor{yellow!20}{\phantom{0}27.27} & \cellcolor{yellow!20}{75.00} & \cellcolor{yellow!20}{\phantom{0}3.16}  & \cellcolor{yellow!20}{0.44}  & 
         \cellcolor{orange!20}{\phantom{00}0.00} & \cellcolor{orange!20}{\phantom{0}27.27} & \cellcolor{orange!20}{100.00} & \cellcolor{orange!20}{\phantom{0}3.55} & \cellcolor{orange!20}{0.32} & \phantom{0}21.21 \\
         Qwen-VL-Max & \cellcolor{green!20}{\phantom{0}18.18} & \cellcolor{green!20}{42.64} & \cellcolor{green!20}{11.36} & \cellcolor{green!20}{0.05} &
         \cellcolor{yellow!20}{\phantom{00}0.00} & \cellcolor{yellow!20}{\phantom{0}27.27} & \cellcolor{yellow!20}{75.00} & \cellcolor{yellow!20}{\phantom{0}3.51}  & \cellcolor{yellow!20}{0.15}  &  
         \cellcolor{orange!20}{\phantom{00}9.52} & \cellcolor{orange!20}{\phantom{0}18.18} & \cellcolor{orange!20}{\phantom{0}94.18} & \cellcolor{orange!20}{\phantom{0}2.72} & \cellcolor{orange!20}{0.31} & \phantom{00}9.23 \\
         Phi-3-vision-128k & \cellcolor{green!20}{\phantom{00}0.00} & \cellcolor{green!20}{50.00} & \cellcolor{green!20}{\phantom{0}0.00} & \cellcolor{green!20}{0.01}& \cellcolor{yellow!20}{\phantom{00}0.00} & \cellcolor{yellow!20}{\phantom{00}0.00} & \cellcolor{yellow!20}{75.00} & \cellcolor{yellow!20}{\phantom{0}0.00} & \cellcolor{yellow!20}{0.02}  &  \cellcolor{orange!20}{\phantom{00}0.00} & \cellcolor{orange!20}{\phantom{0}0.00} & \cellcolor{orange!20}{100.00} & \cellcolor{orange!20}{\phantom{0}0.00} & \cellcolor{orange!20}{0.01} & \phantom{00}0.00\\
         % GLM-4v Flash & \cellcolor{green!20}{\phantom{00}0.00} & \cellcolor{green!20}{50.00} & \cellcolor{green!20}{\phantom{0}0.00} & \cellcolor{green!20}{0.00}& \cellcolor{yellow!20}{\phantom{00}0.00} & \cellcolor{yellow!20}{\phantom{00}0.00} & \cellcolor{yellow!20}{75.00} & \cellcolor{yellow!20}{\phantom{0}0.00} & \cellcolor{yellow!20}{0.00}  &  \cellcolor{orange!20}{\phantom{00}0.00} & \cellcolor{orange!20}{\phantom{0}0.00} & \cellcolor{orange!20}{100.00} & \cellcolor{orange!20}{\phantom{0}0.00} & \cellcolor{orange!20}{0.00} & \phantom{00}0.00\\
         \bottomrule 
    \end{tabular}
    }    
    \vspace{-8pt}
    \caption{Results of standard single room setting. Prop: Prop Gain; Steps: average steps used to complete the game; Grab SR: the precision of grabbing; Grab Ratio: the portion of grabbing actions regarding the total consumed steps. Note that Difficulty-1 requires no prop, and the prop gain is therefore omitted for this setting. The max allowed steps are 50, 75, 100 for Difficulty-1, -2, -3 respectively. The best score of each metrics is \textbf{bolded} and the second is \underline{underlined}.}
    \label{tab:rst}
    % \vspace{-6pt}
\end{table*}

\begin{table*}[t]
    \centering
    \resizebox{\textwidth}{!}{
    \begin{tabular}{l|cccc|ccccc|ccccc}
        \toprule 
         \multirow{3}{*}{Models} & \multicolumn{4}{c|}{Difficulty-1 \& Difficulty-1} & \multicolumn{5}{c|}{Difficulty-1 \& Difficulty-2} & \multicolumn{5}{c}{Difficulty-2 \& Difficulty-2} \\ \cmidrule(lr){2-5} \cmidrule(lr){6-10} \cmidrule(lr){11-15} 
         & ER(\%)$\uparrow$ & Steps$\downarrow$ & {\begin{tabular}[c]{@{}c@{}}Grab \\SR (\%)$\uparrow$\end{tabular}} & {\begin{tabular}[c]{@{}c@{}}Grab\\Ratio\end{tabular}} & 
        ER(\%)$\uparrow$ & {\begin{tabular}[c]{@{}c@{}}Prop\\(\%)$\uparrow$\end{tabular}} & Steps$\downarrow$ &  {\begin{tabular}[c]{@{}c@{}}Grab \\SR (\%)$\uparrow$\end{tabular}} & {\begin{tabular}[c]{@{}c@{}}Grab\\Ratio\end{tabular}} & 
        ER(\%)$\uparrow$ & {\begin{tabular}[c]{@{}c@{}}Prop\\(\%)$\uparrow$\end{tabular}} & Steps$\downarrow$ &  {\begin{tabular}[c]{@{}c@{}}Grab \\SR (\%)$\uparrow$\end{tabular}} & {\begin{tabular}[c]{@{}c@{}}Grab\\Ratio\end{tabular}}  \\
         \midrule 
         GPT-4o & \textbf{75.00} & 35.50  & 34.25  & 0.32 & \textbf{90.00} & \textbf{100.00} & \textbf{34.90} & 35.52 & 0.31 & \textbf{70.00} & \textbf{80.00}  &  \textbf{39.50} & \textbf{42.32} & 0.37 \\
         Gemini-1.5-pro & 22.22 & 40.22 & 22.89 & 0.38 & 40.00 & 50.00 & 56.60 & 16.79 & 0.05 & 60.00 & \textbf{80.00} & 60.00 & 22.71 & 0.34\\
         Llama-3.2-11b-vision & 55.56 & \textbf{31.00} & \textbf{36.25} & 0.36 & 10.00 & 60.00 & 66.40 & 4.40 & 0.40 & 10.00 & 40.00 & 76.80 & 27.00 & 0.19\\
         Claude 3.5 Sonnet & 22.22 & 45.22 & 10.62 & 0.08 & 20.00 & 20.00 & 71.90 & 6.75 & 0.09 & 10.00 & 10.00 & 80.00 & 23.20 & 0.06 \\
         Qwen-VL-max & 22.22 & 40.33 & 12.96 & 0.16 & 30.00 & 50.00 & 57.70 & \textbf{42.30} & 0.28 & 0.00 & 10.00 & 80.00 & 23.66 & 0.32 \\
         \bottomrule 
    \end{tabular}
    }
    \vspace{-8pt}
    \caption{Performance on multi-room setting for different room scenes. 
    To assist in the more challenging setting, we provide models with a full successful escape path from Room 1 (9 steps) for self-reflection when they try to unlock Room 2. Hence, the Prop Gain (Prop (\%)) in the results refers only to Room 2. 
    Further challenges of escaping from the very beginning of multi-room setting are discussed in Supplementary Material~\ref{app:full_multi}.
    }    
    \label{tab:rst_multiroom}
    \vspace{-12pt}
\end{table*}

% \vspace{-9pt}
\paragraph{Metrics of MM-Escape}
To comprehensively evaluate complex multimodal reasoning ability, we propose a set of metrics for the intermediate process, alongside an indicator for final task completion.
For the room escape task, we use average escape rate (ER) as the indicator of game completion, and design five metrics for measuring intermediate interactions, including prop gain, average steps, grab count, grab success rate, and grab ratio. Denoting the total steps as $S$, amount of succeeded grabbing action as $N_{grab}^{TP}$, grab success rate as $\text{GSR}$, grab ratio as $\text{R}_{grab}$ we have,
\vspace{-6pt}
\begin{equation}
        \text{Prop Gain} = \frac{N_{grab}^{TP}}{\sum \text{Prop count}},  
\end{equation}
\vspace{-6pt}
\begin{equation}
        \text{GSR} = \frac{N_{grab}^{TP}}{\sum \text{Grabbing actions}}, 
\end{equation}
\begin{equation}
        \text{R}_{grab} = \frac{\sum \text{Grabbing actions}}{S}, 
\end{equation}

The debriefing task is only applied to cases where models successfully escaped as a bonus, to further investigate the multimodal understanding and reasoning process. Because models that successfully complete the game tend to achieve high scores regarding metrics mentioned above. We employ large language models as evaluator for this task, assessing the consistency ($[0,1]$) of reconstructed stories with the groundtruth stories. 
\section{Experiments}

\paragraph{Evaluation Setups.}
We investigate both open-source models and propriety models of different scales, and evaluate their multimodal reasoning ability in level-wise approaches as outlined in Section~\ref{sec:game_setting}. Specifically, we employ both single room and multi-room settings. The multi-room setting can be viewed as combinations of two single rooms, with the second room featuring two doors—one for entry and one for exit. For robust comparison, we set the temperature to 0 to eliminate token decoding diversity. The prompt templates used for evaluation are detailed in Supplementary Material Sec.~\ref{app:prompt}. Since the game can grow infinitely, we limit the maximum steps for each difficulties as 50, 75, and 100 respectively, for quantitative measurements. We slightly increase the max step to 80 for the most challenging multi-room setting. Since the proposed room escape task requires models to fully utilize past trajectories to gain a comprehensive understanding of the room environment, we investigate models supporting long input contexts, at least 128k, as shown in Table~\ref{tab:rst}. Proprietary models include, GPT-4o\footnote{\url{https://platform.openai.com/docs/models/gpt-4o}}, Gemini-1.5-pro\footnote{\url{https://ai.google.dev/gemini-api/docs/models\#gemini-1.5-pro}}, Claude 3.5 Sonnet\footnote{\url{https://www.anthropic.com/news/claude-3-5-sonnet}} and Doubao 1.5 Pro\footnote{\url{https://www.volcengine.com/docs/82379/1330310\#doubao-1-5-vision}}, open-source models include Phi-3-vision, Llama-3.2-11b and Qwen2.5-VL.

\begin{figure*}
    \centering
    \includegraphics[width=1.\linewidth]{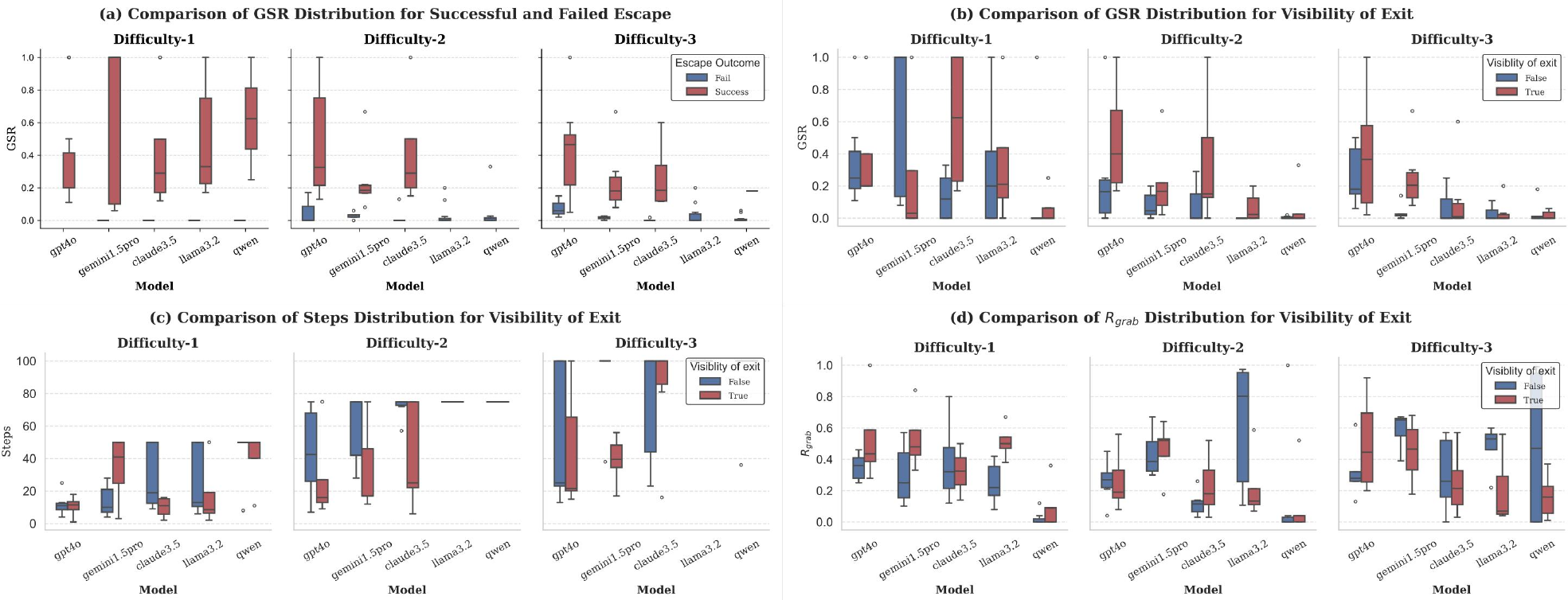}
    \vspace{-21pt}
    \caption{Analysis on the grabbing behaviors with respect to the escape rate and the visibility of the exit at initial location.}
    \label{fig:analysis-grab}
    \vspace{-10pt}
\end{figure*}

\vspace{-11pt}
\paragraph{Main Results of Game Completion}
Results of standard single room setting, along with manual evaluation are reported in Table~\ref{tab:rst}. See Supplementary Material Sec.~\ref{app:human} for details of human evaluation.

The results indicate that model performances falls significantly behind human capabilities across all metrics, highlighting the remarkable gap between current multimodal reasoning abilities and human-level capabilities.
Among the models, GPT-4o demonstrates the strongest overall performance across all difficulty levels, achieving the highest average escape rate (81.36\%), followed by Gemini-1.5-pro (61.06\%). Other models limited success in task completion, except for Phi-3-vision-128k, which fails considering the task completion. Claude 3.5 Sonnet and Llama-3.2-11b-vision achieve moderate ER, but their performance drops significantly in Difficulty-2 and -3, suggesting weaknesses in multi-hop multimodal reasoning.

For the intermediate process, GPT-4o efficiently completes tasks with relatively fewer steps while maintaining a high grabbing success rate. Notably, Gemini-1.5-pro achieves a lower escape rate (ER) in Difficulty-2 compared to GPT-4o, yet it presents the highest Prop Gain, suggesting a strong visual search ability. For models do not survive the task completion, we can still measure their multimodal reasoning ability in terms of Prop Gain, GSR and Grab Ratio. Llama-3.2-11b-vision and Qwen-VL-Max receive identical Prop Gain, while the later presents higher GSR with a lower Grab Ratio. This implies that Qwen-VL-Max can precisely identify scenes with useful props, while Llama-3.2 is likely to adopt a greedy policy to increase grab actions so as to find more props, which is often unintentionally.

For the more challenging multi-room setting, we can derive similar conclusions. Additionally, we notice that by providing a successful path of the first room helps the model to better conduct multimodal reasoning in our task for most models, but the improvements present in different forms. For example, GPT-4o achieves an ER of 90\% for Difficulty-2 when bootstrapped by a first room of Difficulty-1, and prop gain of Gemini and Llama 3.2 is improved in Difficulty-2 \& -2 combination compared with the setting of a single room of Difficulty-2.

\begin{table}[]
    \centering
    \scriptsize
    % \resizebox{0.5\textwidth}{!}{
    \begin{tabular}{ll|cc|ccc}
    \toprule
     & Models & \multicolumn{2}{c|}{Difficulty-2} & \multicolumn{3}{c}{Difficulty-3} \\
     \cmidrule(lr){3-4} \cmidrule(lr){5-7}
    & & \#Key & \#Exit & \#PW & \#Key & \#Exit  \\
    \midrule
    \multirow{3}{*}{\rotatebox{90}{\textit{step}}} & GPT & \textbf{16.25} & \textbf{39.10} & 11.80 & \textbf{17.23} & \textbf{50.40} \\
    & Gemini & 16.50 & 50.90 & 20.67 & 38.13 & 65.70 \\
    &  Claude & 59.60  & 62.60 & \textbf{5.25} & 20.67 & 84.80 \\
    \midrule
    \multirow{3}{*}{\rotatebox{90}{\textit{cost}}} 
    & GPT & 0.68 & 0.40 & 0.36 & 0.28 & 0.40\\
    & Gemini & 0.62 & 0.63 & 0.49 & 0.29 & 0.31 \\
    &  Claude  &0.91 & 0.23 & 0.66 & 0.50 & 0.46  \\
    \bottomrule
    \end{tabular}
    % }
    \vspace{-6pt}
    \caption{Steps taken to get key props and to open the door, denoted as \textit{step}. And the ratio of cost of each corresponding stages, denoted as \textit{cost}, which is the proportion of corresponding steps out of the total steps. \#PW: \textit{step}/\textit{cost} to obtain passwords from the beginning. \#Key: \textit{step}/\textit{cost} to obtain keys from the beginning (Difficulty-2), and to obtain keys after the acquisition of password (Difficulty-3). \#Exit: \textit{step}/\textit{cost} to escape after the acquisition of the last key prop.}
    \label{tab:grab_ana}
    \vspace{-12pt}
\end{table}

\section{Analysis and Discussions} \label{label:ana}	

\paragraph{Takeaway Observations}
\begin{itemize}
	\item \textit{Distinct Human-Like Behavioral Patterns}: Models exhibit unique behaviors in room escape task. Gemini tends to remain in a fixed location, typically the starting point, scanning its surroundings before taking action. In contrast, GPT-4o first observes a wider range for a global understanding of the environment.
	\item \textit{Diverse Observation Strategies}: Some models deviate from a conventional front-facing view. Gemini frequently employs a downward-facing perspective to inspect objects like tables and chairs, whereas GPT-4o predominantly relies on a front-facing view.
    \item \textit{Common failure modes}: We observe some common reasons leading to unsuccessful escape, such as movement failures and interaction challenges. Movement failures vary across models, including repetitive trajectories (GPT-4o) and getting trapped (Gemini and Claude). Interaction challenges include action combination failure (Phi-3 and Qwen-VL) and imprecise object specifying. 
	\item \textit{Adherence to Long-Term Goals}: Most models remain goal-oriented, focusing on locating the exit and clues or props that could assist in escaping. However, Phi-3 consistently fails to generate valid object interactions and does not recognize doors, unlike other models, which engage with doors upon detection.
\end{itemize}
% \vspace{-6pt}
Please refer to Supplementary Material Sec.~\ref{app:takeaway} for full version of overall observations. In the following of this section, we will further discuss model performance concerning the escape process, including the escaping path, the consistency of sub-goals, and post-game debriefing, and also provide cases to demonstrate the extensibility of our EscapeCraft.

\subsection{Analysis of Entire Path}

\begin{figure*}[t]
    \centering
    \includegraphics[width=1.\linewidth]{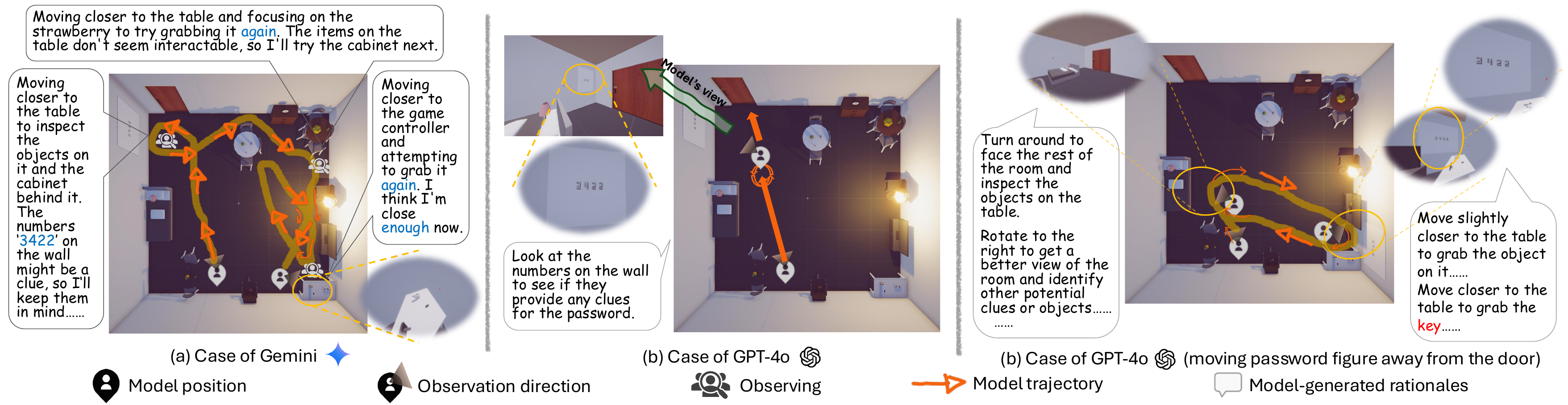}
    \vspace{-18pt}
    \caption{The extended scenario where the required password is displayed via a numerical pattern on the wall, rather than explicitly written on notes. GPT-4o completes reading it at once and exits within five steps, while Gemini struggles to repetitively search the room. Moving the pattern away from the door further challenges GPT-4o, leading to a failure of escaping.}
    \label{fig:new_room}
    \vspace{-12pt}
\end{figure*}

We investigate three three key questions in this section: Q1) How many steps required to obtain props?
Q2) How many steps are needed to exit the room after acquiring the core prop (key or password)?
Q3) What is the relationship between grab success rate and escape outcome for each test?

For Q1, GPT-4o presents a significant advantage in step counts required to obtain the key followed by Gemini as shown in Table~\ref{tab:grab_ana}. While Claude requires fewer steps to find props in Difficulty-3, this comes at the cost of a significant decrease in escape rate. The superior performance in locating and obtaining the core props can be attributed to better understanding of task objectives and the holistic environment, and its enhanced reasoning abilities in this context.
For Q2, Gemini can locate and acquire the key at a lower cost in difficulty-2, but GPT-4o outperforms in Difficulty-3, which is more complex. GPT-4o benefits from its prior memory and understanding of the room environment, gained in the process of obtaining key props, which aids it to efficiently locate the exit and escape with fewer steps compared to other models.
For Q3, escape success is positively correlated with grab success rate (GSR), as shown in Figure~\ref{fig:analysis-grab}(a). Higher GSR implies that models have experienced more successful interactions with the environment, potentially indicating a clearer understanding of the overall environment and ultimate goals for our task. While GSR declines with difficulty, the scores of GPT-4o and Claude 3.5 remain relatively stable compared to others, with less variation in grabbing behavior and GSR across difficulties. The low GSR of Qwen in difficulty-2 and -3 can be partly caused by the ineffective perception of the environment, inferior reasoning and interacting decision in this complex tasks, while the low GSR of Llama 3.2 is limited by its input registration of only one image at a time.

Please refer to Supplementary Materials Sec.~\ref{app:grab} for detailed discussion, and Sec.~\ref{app:moving} for additional examination about the moving distance and arrangement of the room.

\subsection{The Extensibility of EscapeCraft}\label{sec:case_study}
% We provide video demos of our experiments with standard settings in Supplementary Materials, and provide an extended case study in this section.
We provide an extended case study in this section, and also discuss the fully autonomous version of multi-room setting in Supplementary Materials Sec.~\ref{app:full_multi}, and an additional customizations of escaping path in Supplementary Materials~\ref{app:reverse}.

We introduced an extended scenario where the required password is displayed via a numerical pattern on the wall, rather than explicitly written on notes, as shown in Figure~\ref{fig:new_room}. Models should recognize the pattern on the wall (password) and infer its relevance to the door. When pattern appears near the door, GPT-4o quickly identifies it and exit in the following five steps, while Gemini, despite seeing the pattern, failed to recognize it as the password and instead searched the room exhaustively repeatedly.

We further move the pattern away from the location of the door, and observe unchanged behaviors of Gemini. However, GPT-4o performs differently, by repeatedly moving between the bed and the wall without recognizing the password. It also failed to interact with the door until the game stops by the max allowance, revealing limitations in its long-term reasoning and spatial reasoning.

\subsection{Analysis of Intermediate Consistency}

\begin{table}
    \centering
    % \vspace{-12pt}
    \resizebox{\columnwidth}{!}{
    \begin{tabular}{c|ccc}
    \toprule
         Models & Claude 3.5 & GPT-4o & Gemini 1.5 pro \\
         % & Gemini 2.0 flash\\ 
         \midrule
         Scores: $C_{IO} (GSR)$ & 21.74 (16.21) & 26.51 (31.36) & 26.64 (10.43) \\
         % & 26.26 (-) \\
    \bottomrule
    \end{tabular}
    }
    \vspace{-9pt}
    \caption{Consistency compared with GSR on Difficulty-3.}
    \label{tab:consis}
    \vspace{-12pt}
\end{table}

During experiments, we observe that some successful interactions are completed by accident instead of a proper reasoning process. For example, a model aims at a TV but obtaining a key instead. To further analyze the reasoning process concerning successful sub-goals (i.e. successful grabbing), we introduce an additional metric, \textit{intent-outcome consistency}, denoted by $C_{IO}$, ranging [0,1]. This metric quantitatively assesses the intermediate process of achieving sub-goals. Specifically, it measures if model generated rationale (\textit{intent}) aligns with actual environment feedback (\textit{outcome}). We use GPT-4o for auto-evaluation with prompt template provided in Supplementary Materials~\ref{app:consis_prompt}, where score 0 implies accidental success, and 1 reflects true valid reasoning. We calculate scores of Difficulty-3, as it involves more steps which can better reflect the intermediate consistency, and report some of the results in Table~\ref{tab:consis}. GPT-4o achieves only 26.51\% consistency, despite a 31.36\% of GSR, implying many completed sub-goals are accidental. And so do other models, suggesting common accidental completion of sub-goals, indicating a need to improve the consistency between reasoning and perception capabilities.

\subsection{Analysis on Post-game Debriefing}

The post-game debriefing task requires models to recall their escape process and obtained clues, and reconstruct the whole stories. As successful escape is necessary for post-game debriefing, we only evaluate models with high success rate, that is, GPT-4o and Gemini-1.5-pro. Results show that both models fall short of ability of retelling the stories. Models pay strong attention to the processes which are directly related to the completion of room escaping, such as password acquisition. Meanwhile, they ignore background stories which are less important but also helpful for escaping. For the reason of limited model abilities, this may be a effective strategy to complete tasks. However, with the enhancement of model abilities in the future, it is necessary to improve model ability of memorization of background information. For the experiment results, please refer to Supplementary Material Sec.~\ref{app:debriefing}.
% \section{Future Potentials}

% can build more powerful MLLMs or agent systems with our env. 

% For the steps, the vanilla setting is to directly use all the history steps. To avoid redundant information and to fit the super-long context into the permitted length of the input window, we implement a key-step setting where only steps with successful interaction between env and the player will be recorded in the game history.

\section{Conclusions}

In this paper, we introduce MM-Escape, an extensible benchmark for investigating multimodal reasoning, inspired by real-world escape games. 
We also develop EscapeCraft that enables models to engage in free-form exploration for assessing multimodal reasoning, to construction our benchmark.
We find that MLLMs can successfully complete the simplest level of tasks, and some models even exhibiting human-like behaviors and strategies. However, performance dramatically drops as task difficulty increases, while human testers consistently succeed. MM-Escape reveals distinct failure modes across models, such as repetitive trajectories without adaptive exploration, trapped in corners without good spatial awareness. 
We hope our work sheds light on new challenges, and uncovers potential improvements for MLLMs.

\vspace{-6pt}
\section*{Contributions}
\vspace{-6pt}
\textbf{Ziyue Wang}: Design of the escape process and post-game debriefing, implementation of EscapeCraft, all reported experiments. Paper writing: all sections and figures.

\noindent \textbf{Yurui Dong}: Design of the escape process and post-game debriefing, construction of 3D environment, all engineering and coding works. Paper writing: method sections and appendix sections, figures of case study.

\noindent \textbf{Fuwen Luo}: Design of the escape process, design and experiments of post-game debriefing. Paper writing: related works, post-game debriefing experiments and analysis.

\noindent \textbf{Minyuan Ruan}: Implementation of EscapeCraft, scene generation, construction of homepage. Paper writing: analysis, human evaluations, figures of analysis.

\noindent \textbf{Zhili Cheng}: Construction of 3D environment, design of room escape process. 

\noindent \textbf{Chi Chen}: Design of room escape process and post-game debriefing, support on experiments.
 
\noindent \textbf{Peng Li}: Project supervision, advising of all designs, engineering, experiments, and paper writing.

\noindent \textbf{Yang Liu}: Project supervision, advising of all designs, engineering, experiments, and paper writing. \phantom{0000}
{
    \small
    \vspace{-12pt}
    \bibliographystyle{ieeenat_fullname}
    \bibliography{main}
}

% WARNING: do not forget to delete the supplementary pages from your submission 
\appendix
\clearpage
\setcounter{page}{1}

\maketitlesupplementary

\section{Full Takeaway Observations} \label{app:takeaway}
\begin{itemize}
    \item \textit{Distinct human-like behavioral patterns}: There are distinct behavioral patterns among different models in the room escape task. GPT-4o and Gemini exhibit the most human-like behavior, but the patterns differ from each other. For example, GPT-4o demonstrating stable viewpoint control and deliberate movements. At beginning steps, it tends to walk around with large distances and observe the surroundings in a wide range. While Gemini tends to look around before taking further actions to move to other locations. Some other models tend to shift their perspectives arbitrarily, leading to inefficient exploration. Gemini frequently gets stuck, struggling with spatial navigation, while Phi-3 exhibits a tendency to continuously rotate in place with minimal actual movement, hindering effective exploration.  
    \item \textit{Robustness towards system prompts}: Most of the models are faithful to the system instruction. They are aware of the ultimate target, to locate the door and make a way out, and are not easily distracted by findings during the exploration, except for Phi-3 who always fail to generate required actions with valid and interactable items.
    \item \textit{Common failure modes}: However, significant limitations persist across all model, including GPT-4o sometimes. A common failure mode is inaccurate object positioning within the field of view. Models often fail to center the target object, which is indicated by a guiding red dot in our environment, precisely suggesting the objects to interact with. This often leads to unsuccessful interactions such as grabbing or entering for password. Furthermore, some models struggle with tool utilization, particularly in cases requiring abstract reasoning, such as correctly applying a password to unlock a door. 
\end{itemize}

\section{Human Evaluation}\label{app:human}
\begin{table*}[t]
    \centering
    \small
    % \resizebox{\textwidth}{!}{
    \begin{tabular}{c|c c c c c c c c c c c}
    \noalign{\vskip 2pt} 
        \hline
        \noalign{\vskip 2pt}
        \multirow{2}{*}{\textbf{Metrics}} & \multicolumn{11}{c}{\textbf{Scene}} \\ 
        \noalign{\vskip 2pt} 
        \cline{2-12}  
        \noalign{\vskip 2pt} 
        & 0 & 1 & 2 & 3 & 4 & 5 & 6 & 7 & 8 & 9 & 10 \\  
        \noalign{\vskip 2pt} 
        \hline
        \rowcolor{gray!10}
        \multicolumn{12}{c}{\textbf{\textit{\small Difficulty-1}}}\\   
        \noalign{\vskip 2pt}
        Steps  & 10 & 3 & 7 & 7 & 5 & 5 & 6 & 3 & 7 & 4 & 6\\ 
        Prop Gain(\%)    & 100.00 & 100.00 & 100.00 & 100.00 & 100.00 & 100.00 & 100.00 & 100.00 & 100.00 & 100.00 & 100.00 \\ 
        Grab Count   & 2      & 1      & 1      & 1      & 1      & 1      & 1      & 1      & 1      & 1      & 1 \\ 
        Grab Success(\%) & 50.00  & 100.00 & 100.00 & 100.00 & 100.00 & 100.00 & 100.00 & 100.00 & 100.00 & 100.00 & 100.00 \\ 
        \noalign{\vskip 2pt}
        \hline
        \rowcolor{gray!10}
        \multicolumn{12}{c}{\textbf{\textit{\small Difficulty-2}}}\\   
        \noalign{\vskip 2pt}
        Steps   & 23 & 17 & 10 & 8 & 9 & 13 & 15 & 8 & 16 & 20 & 11 \\ 
        Prop Gain(\%)    & 100.00 & 100.00 & 100.00 & 100.00 & 100.00 & 100.00 & 100.00 & 100.00 & 100.00 & 100.00 & 100.00 \\ 
        Grab Count   & 3      & 3      & 2      & 2      & 2      & 3      & 3      & 2      & 3      & 3      & 2 \\ 
        Grab Success(\%) & 66.67  & 66.67  & 100.00 & 100.00 & 100.00 & 66.67  & 66.67  & 100.00 & 66.67  & 66.67  & 100.00 \\ 
        \noalign{\vskip 2pt}
        \hline
        \rowcolor{gray!10}
        \multicolumn{12}{c}{\textbf{\textit{\small Difficulty-3 (note-key)}}}\\   
        \noalign{\vskip 2pt}
        Steps     & 22 & 20 & 21 & 17 & 23 & 27 & 18 & 16 & 22 & 23 & 27\\ 
        Prop Gain(\%)  & 100.00 & 100.00 & 100.00 & 100.00 & 100.00 & 100.00 & 100.00 & 100.00 & 100.00 & 100.00 & 100.00 \\ 
        Grab Count        & 4      & 5      & 4      & 4      & 4      & 5      & 3      & 3      & 4      & 4      & 5 \\ 
        Grab Success(\%) & 75     & 60     & 75     & 75     & 75     & 60     & 100    & 100    & 75     & 75     & 60 \\ 
        \noalign{\vskip 2pt}
        \hline
        \rowcolor{gray!10}
        \multicolumn{12}{c}{\textbf{\textit{\small Difficulty-3 (key-note)}}}\\   
        \noalign{\vskip 2pt}
        Steps& - &   22 & 21 & 19 & 18 & 20 & 24 & 16 & 27 & 17 & 18\\ 
        Prop Gain(\%)  & - & 100.00 & 100.00 & 100.00 & 100.00 & 100.00 & 100.00 & 100.00 & 100.00 & 100.00 & 100.00 \\ 
        Grab Count  & -  & 4      & 4      & 4      & 5      & 5     & 6      & 4      & 5      & 4      & 4    \\ 
        Grab Success(\%) &- & 75     & 75     & 75     & 60     & 60     & 50     & 75    & 60    & 75     & 75  \\ 
        \noalign{\vskip 2pt}
        \hline
    \end{tabular}
    % }
    \caption{Detailed results for various levels.Since humans completed all escape tasks in the evaluation, the escape rate is 100\% and not reflected above.}
    \label{tab:hunam_results}
\end{table*}

We conduct manual evaluation on MM-Escape, and report detailed results in Table~\ref{tab:hunam_results}. Human participants in EscapeCraft exhibit a clear understanding of how to complete tasks efficiently. By observing objects in the environment, they can make reasonable judgments about which items to pick up, leading to a higher success rate in effective item acquisition and usage. Additionally, when unable to open doors or interactable objects, humans are more adept at promptly shifting their approach to seek alternative clues in the environment rather than getting stuck. 
In terms of spatial awareness, they demonstrate a strong ability to perceive the relative positions of objects, enabling logically reasonable, smoother and more coordinated exploratory actions. Across Difficulty-1 to 3, human participants consistently identify all necessary items with less interaction attempts compared with MLLMs, and successfully complete all the tasks within a limited number of steps.

\section{Construction Details}

\subsection{Environment Construction}
\label{app:env_cons}
\subsubsection{Room Generation}
We adopted the automated 3D room generation method ProcTHOR~\cite{deitke2022}, with additional improvements to enhance its flexibility and applicability regarding diverse type of scenes. Following Procthor, we generate 3D environments that can simulate diverse real-world scenes, such as bedrooms, living rooms, and offices by maintaining collections of typical objects that are common in different scenes. For instance, desks in offices, workbenches in laboratories, and other representative objects of corresponding scenes. We enable automatic creation of 3D rooms from the collections of each scenes, ensuring that the generated rooms accurately reflect their respective environments.

We use a configuration file to generate each room, specifying the items along with required styles, positions, sizes, and interactivity. This enables precise control over the placement of prop objects, ensuring that they are arranged in a manner aligning with real-world expectations on spatial arrangement.

Benefits of the Automated 3D Room Generation include:
\begin{itemize}

    \item Diversity and Complexity: By automatically generating a variety of 3D rooms, we can provide the model with diverse environments, ensuring that it is capable of handling various layouts, objects, and puzzle elements. This diversity is critical in assessing the model’s ability to reason in different scenarios, evaluating its performance when confronted with unknown and complex situations. Moreover, the ability to create different configurations on the fly means the model will not be limited to predefined environments, which helps to prevent overfitting to specific room layouts.

    \item Enhanced Realism: Unlike manually designed fixed scenes, automatically generated 3D environments can simulate more natural and irregular spatial layouts. This is essential for training and evaluating agents on spatial reasoning, pathfinding, and interaction skills. By incorporating a wide range of room designs, we create more realistic scenarios in which agents must navigate, interact with objects, and solve problems, similar to real-world challenges.

    \item Efficiency: The automated generation of 3D rooms significantly improves the efficiency of the testing process. Whether for debugging model performance or conducting large-scale evaluations, the ability to generate various environments quickly eliminates the time-consuming and tedious process of manually creating scenes. This allows for faster iteration and more comprehensive testing without the bottleneck of scene creation.

    \item Evaluation Robustness: In escape room-style games, the diversity of room layouts and puzzles directly influences the game's difficulty. By automating the scene generation process, we enable the model to train and be evaluated in a wide array of environments, which helps enhance its robustness. This diversity allows the model to develop better strategies for handling new and unexpected challenges, improving its ability to generalize across different scenarios.

    \item Prevention of Cheating and Overfitting: One of the major advantages of generating an infinite variety of scenes is the prevention of both cheating and overfitting. Fixed testing environments often lead to overfitting, where a model can "learn" to exploit certain patterns or repetitive features of the environment. In contrast, each automatically generated room is unique, with random elements that require the model to demonstrate true problem-solving abilities in previously unseen configurations. This ensures that the model cannot simply memorize the environment but must adapt its strategies to succeed.

    \item Adaptive Adjustment: Another key feature of our approach is the ability to dynamically adjust the agent’s starting position and other environmental variables. This feature allows us to test how the agent performs under different initial conditions, such as varying the agent’s starting location, the distribution of objects, or the complexity of the puzzle. These adjustments enable a more comprehensive assessment of the agent’s performance, providing deeper insights into its ability to adapt and solve problems in diverse situations.

\end{itemize}
The automated 3D room generation framework we developed not only enhances the diversity and realism of testing environments but also optimizes the efficiency of large-scale evaluations. By providing a mechanism for dynamically altering the environment and agent conditions, it offers a more robust and fair evaluation process, ensuring that models are evaluated under realistic, varied, and challenging conditions.

\subsubsection{Action Space}\label{app:action}
In our \textbf{EscapeCraft} environment, the agent is allowed to perform a set of actions that facilitate its interaction with the environment. These actions include moving forward, rotating right, rotating down, looking at specific coordinates, grabbing objects, and interacting with elements in the environment. Each of these actions plays a distinct role in enabling the agent to explore and solve tasks within the escape scenario.
\begin{itemize}

    \item Moving Forward: This action specifies the distance the agent needs to travel along its current heading. The agent's movement is controlled by the distance parameter, which dictates how far it should move in a straight line.

    \item Rotating Right: This action specifies the angle by which the agent should rotate to the right. The agent can adjust its orientation by a specified angular increment, which allows it to navigate through the environment by changing its field of view.

    \item Rotating Down: Similar to rotating right, this action allows the agent to lower its head by a specified angle. This action is crucial for examining objects at different vertical levels, contributing to a more thorough exploration of the environment.

    \item Looking At: The "looking at" action involves orienting the agent’s view towards a specific coordinate within its current field of view. This coordinate is represented in a relative manner, with the center of the field of view denoted as (0.5, 0.5). By specifying the target coordinates, the agent can focus on particular objects or areas of interest in the environment.

    \item Grabbing: The grabbing action indicates if the agent wants to pick up an object or interact with an item within its proximity. This action is typically used when the agent identifies an object that can be picked up or manipulated, allowing it to add that item to its inventory or interact with it to get crucial information.

    \item Interacting: The interaction action is multifaceted and depends on the context of the object the agent is engaging with. Interactions fall into three primary categories:
    \begin{enumerate}

        \item Item Usage: The agent can use items from its inventory by referencing the unique ID of an item, such as a key, tool, or piece of equipment that it has previously obtained. In these cases, the agent specifies the item ID and applies it to relevant objects in the environment (e.g., using a key to unlock a door).
        \item Text Input: Some interactions require the agent to input text, such as a password to unlock a combination lock. These textual inputs are necessary to progress in the environment when dealing with specific security mechanisms.
        \item Read: When the agent wants to know the detailed information of an item in its inventory (e.g., reading the content recorded in a note), it can use this parameter and provide the ID of the corresponding item to the item usage field.
    \end{enumerate}
A special case arises when the agent performs the grab action and leaves the interaction input empty. In this instance, it indicates the agent's intent to pick up an item within the field of view, without specifying a particular item to interact with. This action is used when the agent is trying to collect objects that are relevant to its escape mission.

\end{itemize}

Throughout the agent's exploration, its interactions with the environment yield varying types of feedback. The environment is populated with different types of objects, classified as follows:
\begin{itemize}

    \item Non-Escape Related Props: These are objects within the environment that do not directly contribute to the agent's escape objectives. Interactions with these items provide no information or progress.

    \item Collectible Items: These items can be obtained and added to the agent's inventory, providing critical information and/or utility for the agent’s tasks. Upon collection, the agent gains knowledge of the item’s identity and its associated attributes.

    \item Locked Props: These include objects such as locked doors, chests, or other secured items. When the agent interacts with a locked object in the early stages, it receives a prompt indicating the type of item required to unlock it. Upon obtaining the corresponding item (e.g., a key, a code, or another unlocking mechanism), the agent can use the appropriate item from its inventory to unlock the object by specifying its ID or providing the required input (e.g., entering a password). Once these items are unlocked, the agent will immediately obtain the props contained in them and be informed of the simple information of the items obtained.

\end{itemize}

These interaction dynamics are crucial for the agent’s progression in the environment, as they form the basis for decision-making, object management, and problem-solving. The design of these interactions reflects the need for both exploration and strategy, with the agent needing to acquire, manage, and apply various items in order to navigate and ultimately escape the environment.

\subsection{Data Construction}
\label{app:data_cons}

\subsubsection{Prop Chain}

We proposed a procedural generation approach for constructing game settings tailored to overcome the inherent limitations of current language models, such as restricted context length and reduced reasoning capabilities. To address these constraints, we propose the concept of \textbf{Prop Chain}, a singly linked list that organizes interactive game elements in a sequence, ensuring a coherent flow of gameplay interactions. Each node in the linked list corresponds to a distinct interactive item or action, such as a key, a locked box, or a note with a password. The tail node of the chain signifies the game’s exit point, thereby serving as the conclusion of the sequence. Table~\ref{tab:prop_chain_example} shows the the Prop Chain for the Difficulty-3 Level.

In our implementation of the \textbf{Prop Chain}, we initially focus on a set of fundamental game elements: a key, a locked box (which can only be opened with a key or password), a note (carrying both password and story-related information), and an exit (which is locked and requires either a key or password to access). These components are used to construct a series of interconnected nodes, where each item or action is represented by a node in the chain. The links between the nodes define the relationships between the props and the ways in which they can be obtained or used during the game. For instance, some props may be freely accessible, while others require specific conditions, such as possessing a key to unlock a box, or using a specific password to open the door.

The inter-node relationships can be annotated to represent different interaction modes. For example, a key can be placed within a box, requiring the player to first unlock the box before acquiring the key. Additionally, nodes are allowed to contain multiple conditions. A note revealing the password to the exit are both narrative (to see or infer the textual password) and functional (to open the door).

Each node has an additional show property set to indicate whether the item should appear directly in the scene (for example, a key placed in a box only needs to show the box in the scene, while a key that can be directly obtained independently needs to be shown in the scene), allowing us to determine which props need to be generated in the 3D scene by reading the game settings.

While our initial focus on a limited set of props and interactions, such as the key, locked box, note, and exit, suffices for creating a variety of escape game settings that challenge current language models, the system is highly extensible. The procedural nature of \textbf{Prop Chain} allows for the seamless integration of new props, interactions, and unlocking mechanisms. As such, the framework can easily accommodate additional types of interactive items, more intricate unlock conditions, and customized gameplay mechanics in future iterations. This scalability ensures that the approach remains adaptable to more complex and diverse game scenarios, further enhancing its applicability for testing language models in a variety of settings.

The \textbf{Prop Chain} framework provides a robust and flexible methodology for the procedural generation of game settings. By focusing on a set of core interactive elements and defining their relationships within a linked list structure, we have developed a scalable approach that can evolve to incorporate new game dynamics and meet the increasing demands of future language models.

\begin{table*}
\centering
\begin{tabular}{l|l|lp{6cm}|l}
\toprule
\textbf{ID}       & \textbf{Type}  & \textbf{Unlock Method}      & \textbf{Contents}                          & \textbf{Show} \\ 
\midrule
box\_1            & box            & password (\texttt{password\_1}) & \texttt{key\_1}, \texttt{note\_2}          & true          \\ 
key\_1            & key            &                              & -                                          & false         \\ 
note\_1           & paper          & -                            & password (\texttt{password\_1})            & true          \\ 
note\_2           & paper          & -                            & some story                           & false         \\ 
password\_1       & password       & -                            & -                                          & false         \\ 
\midrule
exit              & exit           & key(\texttt{key\_1})         & -                                          & -             \\
\bottomrule
\end{tabular}
\caption{Representation of the Prop Chain for the Difficulty-3 Level. The level includes a sequence of interactive props where only \texttt{box\_1} and \texttt{note\_1} are visible in the room. The gameplay progression follows a structured sequence: the agent first discovers \texttt{note\_1}, which contains the \texttt{password\_1} needed to unlock \texttt{box\_1}. Inside \texttt{box\_1}, the agent retrieves \texttt{key\_1} and \texttt{note\_2}, the latter of which contains a story element of the game. Finally, the agent uses \texttt{key\_1} to unlock the exit and complete the game.}
\label{tab:prop_chain_example}
\end{table*}

\section{Analysis of Moving Distance}\label{app:moving}
\begin{table}[h]
    \centering\small
    \begin{tabular}{cccccc}
        \toprule
         & GPT & Gemini & Claude & LLaMA & Qwen \\ 
        \midrule
        Correlation & - 0.06  & 0.06  & 0.49 & 0.63 & - 0.48 \\ 
        \bottomrule
    \end{tabular}
    \caption{Correlation between optimal distance and model moving distance.}
    \label{tab:distance-moving}
\end{table}

We calculate the optimal distance required for escape tasks in each scene and compare it with the real distance experienced by the models. Contrary to our expectations, the experienced distance does not exhibit a significant correlation to the distance among key props and the exit within the scene shown in Table~\ref{tab:distance-moving}. This discrepancy may be attributed to the lack of holistic environmental perception of models, which prevents them from further reasoning and planning based on current and ultimate goals, thereby failing to generate an effective and optimal route to complete the task.

\section{Analysis of Grabbing Behaviors} \label{app:grab}
In Figure~\ref{fig:analysis-grab} (b)(c)(d), we analyzed three performance metrics, steps, GRS, and $R_{\text{grab}}$, during the model task completion process under Visibility of Exits at initial locations and orientations . The results indicate that, under common trends, the ability to see the exit from the initial position aids the model in escaping the room with fewer steps.It aligns with our intuition, as the exit, crucially related to the ultimate task goal, plays a significant role in model's visual recognition, reasoning and interaction with the environment to collect information. However, there exist exceptions. For Difficulty-1, some well-performed models still struggle to achieve high GSR and consume more steps despite being able to see the exit from their initial location. They do not interact directly with the exit at the very beginning, and instead choose to rotate around explore the environment for more information and to conduct reasoning and taking action. This is also evident in Difficulty-2 and -3, where these models, after acquiring the key prop, can locate the exit and escape more efficiently, as reflected in better performance in terms of GRS, $R_{grab}$  and steps.

 We further raise three questions for the analysis of the reasoning process during escaping: i) How many steps it costs to obtain props?
ii) How many steps it costs to exit the room after obtaining the core prop (key or password to the door)?
iii) What is the relationship between grab success rate (GSR) and escape outcome for each test?

For question 1, GPT-4o demonstrates a significant advantage in the number of steps required to obtain the key followed by Gemini as shown in Table~\ref{tab:grab_ana}. Although Claude requires fewer average steps to find props in Difficulty-3, this comes at the cost of a significant decrease in escape rate. The superior performance in locating and obtaining the core prop can be attributed to model's better understanding of task objectives and the environment in the escape room, as well as its enhanced reasoning abilities in this context.

For question 2, Gemini is able to locate and acquire the key at lower cost in difficulty-2. But in difficulty-3 which is more complex, GPT-4o performs better. It finds the core prop with fewer steps and its prior memory and understanding of the room environment—gained in the process of obtaining key props—aids it to locate the exit and escape using even fewer steps compared to other models.

For question 3, we observe that escape success is positively correlated with GSR, as shown in Figure~\ref{fig:analysis-grab}(a). A higher Grab SR implies that models have experienced more successful interactions with the environment. It potentially indicates a clearer understanding of the overall environment and ultimate goals within the room escape task, leading to a higher success rate. As difficulty increases, the Grab SR of most models declines, and many of them fail to escape. However, GPT-4o and Claude 3.5 remain relatively stable, with less variation in grabbing behavior and success rate across difficulty settings compared to others. The low success rate of Qwen, and Llama 3.2 11B in difficulty 2 and 3 can be partly attributed to their inability to effectively perceive the environment, reason and make appropriate object interaction choices in more complex tasks.

\begin{table}[t]
    \centering
    % \scriptsize
    \resizebox{0.45\textwidth}{!}{
        \begin{tabular}{@{\hspace{0.1cm}}l|c@{\hspace{0.3cm}}cccc}
        \toprule 
         \multirow{2}{*}[-1.5ex]{Models} & \multicolumn{5}{c}{Difficulty-3-note-key} \\ \cmidrule(lr){2-6} 
         &{\begin{tabular}[c]{@{}c@{}}ER\\(\%)$\uparrow$\end{tabular}} & {\begin{tabular}[c]{@{}c@{}}Prop\\(\%)$\uparrow$\end{tabular}} & Steps$\downarrow$ &  {\begin{tabular}[c]{@{}c@{}}Grab \\SR (\%)$\uparrow$\end{tabular}} & {\begin{tabular}[c]{@{}c@{}}Grab\\Ratio\end{tabular}} \\
         \midrule
         GPT-4o & \textbf{72.73} & \textbf{100.00} & \textbf{47.18}  & \textbf{33.82} & 0.42 \\
         Gemini-1.5-pro &  63.64 & 86.36 & 61.27 & 16.06  & 0.51 \\    
         Claude 3.5 Sonnet & 36.36 & 40.91 & 78.55 & 10.03 & 0.27 \\
        \toprule 
         \multirow{2}{*}[-1.5ex]{Models} & \multicolumn{5}{c}{Difficulty-3-key-note} \\   \cmidrule(lr){2-6} 
         &{\begin{tabular}[c]{@{}c@{}}ER\\(\%)$\uparrow$\end{tabular}} & {\begin{tabular}[c]{@{}c@{}}Prop\\(\%)$\uparrow$\end{tabular}} & Steps$\downarrow$ &  {\begin{tabular}[c]{@{}c@{}}Grab \\SR (\%)$\uparrow$\end{tabular}} & {\begin{tabular}[c]{@{}c@{}}Grab\\Ratio\end{tabular}} \\ 
         \midrule
         GPT-4o & \textbf{70.00}	& \textbf{80.00}	& \textbf{53.20}	& \textbf{28.90} & 0.29\\
         Claude 3.5 Sonnet & 37.50 & 68.75	& 88.14	& 22.05 & 0.15\\
         Gemini-1.5-pro & 30.00	& 60.00	& 87.70	& 4.79 & 0.46 \\
         \bottomrule 
    \end{tabular}
    }    
    \caption{Detailed results of note-key and key-note settings of Difficulty-3.}
    \label{tab:rst_reverse}
\end{table}

\section{Discussion of Fully Autonomous Multi-room Escape}\label{app:full_multi}
We discussed a simplified multi-room setting in Table~\ref{tab:rst_multiroom}. We further study how models behavior in this section. The ER of GPT-4o decreases to only 50\% on average for the settings of applying Difficulty-2 to room 2. The grabbing behaviors also change, where both the Grab SR and Grab Ratio decreases. Similar trends are observed for Gemini and Claude. These indicate that models can learn from a successful escape history. We also note that by setting the two rooms to the same difficulty level further helps models to escape, while different levels do not benefit as expected.

\section{Discussion of Customizing Difficulties}\label{app:reverse}

We enable two different settings of Difficulty-3, a key-note setting and a note-key setting. We observe that human annotators perform equally for both settings (from Table~\ref{tab:hunam_results}), while some models present preferences regarding the key-first and the note-first (i.e. the password-first), as shown in Table~\ref{tab:rst_reverse}. Gemini presents an approaching GPT-4o level results in the note-key setting, while scores the worst in all calculated metrics among the three reported models, presenting a preference towards searching for the note rather than recognizing and interacting with the key. Additionally, Claude scores higher in Grab SR regarding the key-first setting than the note-first setting, potentially indicating a better attention on the key (directly used to unlock the door) than on the note (with clues, implicitly assists with the escape process).

For the multi-room setting, whose results are reported in Table~\ref{tab:rst_multiroom}, we further extend the experiments to a full autonomous scenario to require models to escape both room in order all by themselves. This means, the first room no longer serves as a bootstrapping guidance. We notice a performance drop both in the escape rate and the grabbing behaviors. 

\section{Experiments with Reasoning Models}\label{app:reasoning_model}
Results of recently released reasoning models, such as Claude 3.7 and o1, are reported in Table~\ref{tab:3.7}. We also provide results fo GPT-4o and Claude 3.5 for comparison. 
Notably, o1 and Claude 3.7 attempt fewer grabs but yield higher GSR and Prop Gain, indicating more efficient and intelligent reasoning compared to there previous versions. 

\begin{table}[htbp]
    \centering
    \scriptsize
    % \vspace{-10pt}
    \resizebox{\columnwidth}{!}{
    \begin{tabular}{c|ccccc}
    \toprule
         Models & SR & Prop & Steps & GSR & Grab Ratio\\
         \midrule
         GPT-4o & 72.73 & 81.82 & \textbf{36.73} & 36.73 & 0.26\\
         o1  &  72.73 & \textbf{86.36} & 43.56 & \textbf{39.40} & 0.22 \\
         \midrule
         Claude 3.5 & 45.45 & 54.55 & 57.45 & 20.64 & 0.17\\
         Claude 3.7 & \textbf{54.55} & \textbf{59.09} & \textbf{52.78} & \textbf{38.99} & 0.11 \\
    \bottomrule
    \end{tabular}
    }
    % \vspace{-9pt}
    \caption{Results of reasoning models on Difficulty-2.}
    \label{tab:3.7}
    % \vspace{-10pt}
\end{table}

\section{Results of Post-game Debriefing}\label{app:debriefing}

We choose models with top-2 escaping rate for evaluation, that is, GPT-4o and Gemini-1.5-pro. We first leverage the debriefing prompt, as shown in Table~\ref{tab:story_prompt}, to get recovered stories from models. Then we use GPT-4o to assess the recovered stories according to their similarities to our given groundtruth stories, then assign a score between 0 and 5 to each recovered story. As shown in Table~\ref{tab:result_debriefing}, average scores of both models are low. We further show a case in Table~\ref{tab:case_debriefing}. We can see that models can recall and describe their escape process relatively well, but they do not remember clues about the background stories well, which leads to the failure of restoring the full picture of stories.

\begin{table}[h]
    \centering\small
    \begin{tabular}{ccc}
        \toprule
        Model & GPT-4o & Gemini-1.5-pro \\ 
        \midrule
        Average Score & 0.33  & 0.44 \\ 
        \bottomrule
    \end{tabular}
    \caption{Average scores of models on post-game debriefing task. Scores are between 0 and 5.}
    \label{tab:result_debriefing}
\end{table}

\begin{table}[ht]
    \centering
    \resizebox{\columnwidth}{!}{
    \begin{tabular}{p{\linewidth}}
    \toprule
    \rowcolor{gray!10} \multicolumn{1}{c}{\textit{Groundtruth}} \\
    The story revolves around a secret society that hides within the walls of an ancient mansion. Once you enter, you must follow cryptic clues and navigate through a maze of rituals and secret rooms. The society's true purpose is far darker than its intellectual allure, and the deeper you go, the harder it is to escape. The goal is to uncover the society's mysteries and find a way out before becoming part of its twisted legacy.\\
    \midrule
   \rowcolor{gray!10} \multicolumn{1}{c}{\textit{Recovered Story from GPT-4o}} \\
The room appeared to be a deliberately designed escape challenge, possibly created for entertainment or as a test of problem-solving skills. The presence of the locked box and the notes suggested a sequence of puzzles that needed to be solved to escape.

The sequence of events began with the discovery of Note 1, which provided the numerical code "9926." This code was used to unlock the box on the table, revealing Key 1 and Note 2. The key was then used to unlock the door, allowing for the escape from the room.

The story on Note 2, while not explicitly detailed, might have provided a narrative or thematic context for the escape challenge, perhaps hinting at the creator's intentions or the purpose of the room. The overall experience was one of exploration, deduction, and the satisfaction of solving a series of interconnected puzzles to achieve freedom.\\
    \midrule
   \rowcolor{gray!10} \multicolumn{1}{c}{\textit{Recovered Story from Gemini-1.5-pro}} \\
The room was clearly designed as an escape room challenge. The purpose was to test my problem-solving skills and ability to find and interpret clues. The creator of the challenge remains unknown, but the sterile and impersonal nature of the room suggests a deliberate design choice.  The sequence of events was as follows: I started in the center of the room, explored the cabinet near the door, and then the sofa and stool.  I then found the note with the number "1148".  Realizing the box on the side table was likely password-protected, I used the number from the note to unlock it. Inside, I found the key (key\_1) and a second note (note\_2).  I then used the key to unlock the door and escape the room. The story on note\_2 likely served as a distraction or perhaps a thematic element to the escape room's design, but ultimately, the key to escaping was the numerical code "1148". \\
    \bottomrule
    \end{tabular}
    }
    \caption{A case of post-game debriefing.}
    \label{tab:case_debriefing}
\end{table}

\section{Required Abilities}\label{app:task}
In our \textbf{MM-Escape Benchmark}, the free-form task challenges models with the following abilities: 
% \textcolor{red}{[TODO]}
\begin{itemize}
    \item \textit{Visual search}. The model should have a prior knowledge on the potential locations of useful clues and objects, reason about its current field of view and make decisions to explore the environment efficiently.
    \item \textit{Active perception}. Due to the limited field of view, the model cannot acquire all relevant information at once. It must actively adjust the perspective by moving and rotating its viewpoint, allowing is to perceive from different angles and effectively identify key objects, such as the key, the password and the door.
    \item \textit{Spatial reasoning}. The model is required to deduce spatial depth and objects in the room. It must infer hidden or obscured elements, estimate distances, and determine the most appropriate actions based on the spatial configuration of its surroundings it observed. The model should leverage its understanding of space to make decisions about movement, interaction, and problem-solving, progressively advancing toward the escape objective.
    \item \textit{Target deduction}. The model is only given the objective to escape the room, without explicit instructions on how to achieve it. Hence, it should autonomously infer necessary actions through interaction and exploration, or decompose the ultimate objective into feasible and short-term goals.
    \item \textit{Prop Utilization Capability}. In the escape room environment, the model must maintain awareness of the items in its inventory and determine the appropriate moments to use them. Effective utilization of these props is crucial for navigating the complex environment and achieving the escape objective.
    \item \textit{Long-term reasoning}. Successfully escaping the room requires a prolonged sequence of interactions. The model must analyze and integrate long-form text-image data across multiple key interaction steps to make informed decisions.
\end{itemize}

\section{Prompt Template}\label{app:prompt}
\paragraph{System Prompt} The \textbf{System Prompt} consists of two primary components: the \textbf{Instruction Prompt} and the \textbf{Operation Prompt}. The \textbf{Instruction Prompt} provides the model with contextual information regarding the current environment, its overarching objective, and the approach required to achieve this objective. In contrast, the \textbf{Operation Prompt} delineates, in precise detail, the permissible actions and exploratory methods that the model can employ within the environment. Additionally, it specifies the format and structure of the structured data that the model is expected to generate in response. The complete prompt is shown in table~\ref{tab:system_prompt}.

\begin{table*}[t]
    \centering
    \resizebox{\linewidth}{!}{
    \begin{tabular}{p{\linewidth}}
    \toprule
   \rowcolor{gray!10} \multicolumn{1}{c}{\textit{Instruction Prompt}} \\
You find yourself locked inside a room, and your ultimate goal is to escape the room. i.e. the room escape game.\\\\
You can explore the room, interact with objects, inspect items, and resolve puzzles. If you find doors locked or uninteractable, you probably need to search for keys or passwords to unlock the door when interacting with the environment. You can adopt the following actions to explore the room and interact with objects:\\
    \midrule
   \rowcolor{gray!10} \multicolumn{1}{c}{\textit{Operation Prompt}} \\
   - move\_forward: float, ranged between [-10, 10]. This is the number of meters you want to move forward (negative value means moving backward).\\
- rotate\_right: float, ranged between [-180, 180]. This is the number of degrees you want to turn right (negative value means turn left).\\
- rotate\_down: float, ranged between [-90, 90]. This is the angle you want to adjust your view vertically. Positive value means looking downward, while a negative value means looking upward. Angle 0 means looking straight ahead. \\
- jump: bool, whether you want to jump (can be used together with moving forward), e.g., True represents the action "to jump".\\
- look\_at: list[x: foat, y: float], the range of x and y is [0, 1]. This parameter is the coordinates of the point in the image you want to look at. For reference, the coordinates of the upper left corner of the scene are (0, 0) and the coordinates of the lower right corner are (1, 1). Also to mention that there are on clues on the ceiling.\\
- grab: bool, whether you require to interact with the object located exactly at the center of the scene (marked by a red dot). e.g., to grab the key or to interact with (or open) a box at the center of the scene, set grab=True. The red dot assists in locating the object you require to interact with. You might need to adjust the view or move closer to ensure the red dot is on your target object, through the rotate\_right, rotate\_down, and move\_forward actions. To successfully grab an object, you should center the object via the red dot and be in a certain distance to it. If the grabbing fails, try move closer towards the object. If it fails multiple times at the same position, you should be aware that not all objects are interactable, do not get stucked in uninteractable position.\\
- interactions : dict:\{"use\_item\_id": str, this is the item\_id you require to view or use (when used together with grab=True, it means to use this item to interact with the target object you want to grab, e.g. using item\_id of the key to open the door in the scene), "input": str, this is the message you want to input when interacting with the center object\}.\\
- read: str, this is the item\_id that you want to get detailed information from your bag.\\
- rationale: str, represents the rationale of your action. This should explain your decision-making process and help the agent understand your thinking process.\\\\

You need to return data in the following format of JSON\_string to interact with the scene:\\
\texttt{\{}\\
\hspace{2em}\texttt{       ``move\_forward'': float,}\\
\hspace{2em}\texttt{       ``rotate\_right'': float,}\\
\hspace{2em}\texttt{       ``rotate\_down'': float,}\\
\hspace{2em}\texttt{       ``jump'': bool,}\\
\hspace{2em}\texttt{       ``look\_at'': [x: float, y: float],}\\
\hspace{2em}\texttt{       ``grab'': bool,}\\
\hspace{2em}\texttt{       ``interactions'': \{}\\
\hspace{4em}\texttt{           ``use\_item\_id'': str,}\\
\hspace{4em}\texttt{           ``input'': str}\\
\hspace{2em}\texttt{       \},}\\
\hspace{2em}\texttt{       ``read'': str,}\\
\hspace{2em}\texttt{       ``rationale'': str}\\
\texttt{   \}}\\\\

All of the above operations are optional. If no value is passed in, the interactive operation will not be performed.\\\\

You must follow the above instructions and don't say anything else except for the JSON\_string of operations.\\
    \bottomrule
    \end{tabular}
    }
    \caption{The System Prompt}
    \label{tab:system_prompt}
\end{table*}

\paragraph{Step Prompt} The \textbf{Step Prompt} is designed to provide feedback to the model regarding the outcome of its previous interaction with the environment (if an interaction was attempted). Simultaneously, it informs the model in real-time about the items currently available in its inventory for potential use. Additionally, the prompt serves as a directive, encouraging the model to continue exploration or engage in further interactions.The complete prompt is shown in table~\ref{tab:step_prompt}.

\begin{table*}[h]
    \centering
    \resizebox{\linewidth}{!}{
    \begin{tabular}{p{\linewidth}}
    \toprule
   \rowcolor{gray!10} \multicolumn{1}{c}{\textit{Interaction Result}} \\
\{interaction\_result\} \\
===\\
    \midrule
   \rowcolor{gray!10} \multicolumn{1}{c}{\textit{Inventory}} \\
   
The items in your bag usable include:\\
\{bag\_desc\}\\
===\\
\midrule
   \rowcolor{gray!10} \multicolumn{1}{c}{\textit{Step Prompt}} \\
   Please determine the next action(s) that could help you observe the room or obtain useful tools or clues.\\
If you find yourself stuck in a corner, try turn around by passing rotate\_right.\\
You need to return data in the following format of JSON\_string to interact with the scene and don't say anything else: \\
\texttt{\{}\\
\hspace{2em}\texttt{       ``move\_forward'': float,}\\
\hspace{2em}\texttt{       ``rotate\_right'': float,}\\
\hspace{2em}\texttt{       ``rotate\_down'': float,}\\
\hspace{2em}\texttt{       ``jump'': bool,}\\
\hspace{2em}\texttt{       ``look\_at'': [x: float, y: float],}\\
\hspace{2em}\texttt{       ``grab'': bool,}\\
\hspace{2em}\texttt{       ``interactions'': \{}\\
\hspace{4em}\texttt{           ``use\_item\_id'': str,}\\
\hspace{4em}\texttt{           ``input'': str}\\
\hspace{2em}\texttt{       \},}\\
\hspace{2em}\texttt{       ``read'': str,}\\
\hspace{2em}\texttt{       ``rationale'': str}\\
\texttt{   \}}\\
    \bottomrule
    \end{tabular}
    }
    \caption{The Step Prompt}
    \label{tab:step_prompt}
\end{table*}

\paragraph{Prompt for Consistency Evaluation}\label{app:consis_prompt} The \textbf{Consistency Evaluation Prompt} is designed to assess whether the multimodal agent's reasoning aligns with the actual outcomes of its actions during an escape room interaction. After each interaction, the model is given the agent's internal \emph{rationale}, describing its belief or intended action, and the environment's \emph{response}, which records what actually occurred. The prompt guides the model to judge if the target object mentioned in the rationale matches the object that was truly interacted with, thereby evaluating whether the behavior is intentional or accidental. A special case is defined for successful escapes: the rationale must explicitly or implicitly indicate the agent's goal to exit the room. The model outputs a binary judgment in JSON format, indicating consistency (`1') or inconsistency (`0'). The complete prompt is shown in table~\ref{tab:consistency_prompt}.

\begin{table*}[t]
    \centering
    \resizebox{\linewidth}{!}{
    \begin{tabular}{p{\linewidth}}
    \toprule
    \rowcolor{gray!10} \multicolumn{1}{c}{\textit{Consistency Evaluation Prompt}} \\
You are a reasoning consistency evaluator for a multimodal agent benchmark. Your task is to determine whether the agent's intention (described in the "rationale") matches the actual interaction result (described in the "response").\\\\

Specifically, you are given: \\
- The agent's \textbf{rationale} for why it attempted an interaction, describing its goal or belief about the current environment.\\
- The \textbf{response} from the environment after the interaction, which includes the actual result (e.g., what item was interacted with and what was obtained).\\\\

Your goal is to determine whether the object the agent intended to interact with \textbf{matches} the object that was actually interacted with according to the response.\\\\

If the agent tried to interact with object A (e.g., a microwave), but the response shows interaction with object B (e.g., a box), and B was not the intended target, this is considered an \textbf{inconsistent interaction} (i.e., accidental success or misaligned action).\\\\

\textbf{Special Case:} If the response is ``Escaped successfully!'', you should check whether the agent's rationale explicitly or implicitly indicates the intention to escape (e.g., trying to open the door to leave). If not, treat it as inconsistent.\\\\

You must give your feedback in the following JSON-string format and \textbf{DON'T} say anything else: \\
\texttt{\{} \\
\hspace{2em} \texttt{``Consistency'': 1 | 0} \\
\texttt{\}} \\\\

Where: \\
- \texttt{1} means the rationale and interaction are consistent (i.e., aligned).\\
- \texttt{0} means the interaction appears to be accidental, mismatched, or unintended.\\\\

Respond \textbf{only} with \texttt{Consistency: 1} or \texttt{Consistency: 0}.\\\\

\textbf{---}\\\\

\textbf{Example 1:} \\
rationale: \emph{Moving closer to the microwave to try interacting with it one last time using '1264'. If this doesn't work, I'll need to explore other areas.}\\
response: \emph{You used the correct password to unlock the box... You did not interact with any objects in the last step.}\\
\textbf{Expected output:} \texttt{Consistency: 0} \\\\

\textbf{Example 2:} \\
rationale: \emph{I’ll try using the key I just picked up on the door. Let’s see if I can escape now.}\\
response: \emph{Escaped successfully!}\\
\textbf{Expected output:} \texttt{Consistency: 1} \\\\

\textbf{Example 3:} \\
rationale: \emph{I want to check if the small box has any useful items inside.}\\
response: \emph{You opened the box and found a screwdriver.}\\
\textbf{Expected output:} \texttt{Consistency: 1} \\\\

\textbf{Please score the following interaction:} \\
rationale: \{\texttt{rationale}\} \\
response(s): \{\texttt{response}\} \\
    \bottomrule
    \end{tabular}
    }
    \caption{The Consistency Evaluation Prompt}
    \label{tab:consistency_prompt}
\end{table*}

\paragraph{Debriefing Prompt} The \textbf{Story Recovery Prompt} is used to guide the model to recall and infer the background and story of the entire game based on the interaction records after the model successfully escapes the room. The model is guided to describe the room environment, recall the items that may contain information or clues, and finally piece together the whole story to complete the story recovery. The complete prompt is shown in table~\ref{tab:story_prompt}.

\begin{table*}[ht]
    \centering
    \resizebox{\linewidth}{!}{
    \begin{tabular}{p{\linewidth}}
    \toprule
    \rowcolor{gray!10} \multicolumn{1}{c}{\textit{Story Recovery}} \\
    You have successfully escaped the room. Now, reconstruct the entire story based on the items you discovered during the game and the overall environment you observed. Follow the steps below to guide your recollection and piece together the full narrative.\\
    \midrule
   \rowcolor{gray!10} \multicolumn{1}{c}{\textit{Describe the room environment}} \\
Step 1: Describe the room environment
"Begin by describing the room where you started. What did the room look like? What was the overall atmosphere? Were there any notable features, such as furniture, lighting, or strange objects? Include sensory details like smells, sounds, and the arrangement of the room. This will help set the scene for the story."\\
    \midrule
   \rowcolor{gray!10} \multicolumn{1}{c}{\textit{Recall the items that may contain }} \\
Step 2: Recall the items that may contain information or clues
"Think back to the objects you found throughout the game. What items did you come across? Were any of them unusual or seemed important? These could include physical items like keys, notes, or devices, or even abstract clues like symbols or markings on the wall. Reflect on how each item might have connected to the next step in your escape."   
\\
\midrule
   \rowcolor{gray!10} \multicolumn{1}{c}{\textit{Piece together the whole story}} \\
Step 3: Piece together the whole story
"Now, use the information from the room description and the items you've found to piece together the full story. What was the purpose of the room? Who or what might have created the escape challenge, and why? What was the sequence of events that led you to the escape? Try to connect the dots between the environment, the clues, and the items you encountered, and reconstruct the narrative from start to finish."\\
    \bottomrule
    \end{tabular}
    }
    \caption{The Story Recovery Prompt}
    \label{tab:story_prompt}
\end{table*}

\end{document}